\documentclass[review]{elsarticle}

\usepackage{lineno,hyperref}
\usepackage{float}
\usepackage{amsmath}
\modulolinenumbers[5]
\usepackage{bbm}
\usepackage[dvipsnames]{xcolor}
\usepackage[utf8]{inputenc}  
\usepackage[T1]{fontenc} 
\usepackage{todonotes}
\usepackage{caption}
\usepackage{units}
\usepackage{bm}
\usepackage{tabularx}
\usepackage{algorithm}
\usepackage{algorithmic} 
\usepackage{hhline}
\usepackage{amsfonts}
\usepackage[position=top]{subcaption}
\usepackage{pgfplots}
\usepackage{tikz}
\usepackage[european resistor, european voltage, european current]{circuitikz}
\usetikzlibrary{arrows,shapes,positioning}
\usetikzlibrary{decorations.markings,decorations.pathmorphing,
decorations.pathreplacing}
\usetikzlibrary{calc,patterns,shapes.geometric}

\definecolor{cincinnati-red}{RGB}{190,0,0}

\usepackage{amssymb}

\let\emptyset\varnothing

\begin{document}

\begin{frontmatter}

\title{Variance based sensitivity analysis for Monte Carlo and importance sampling reliability assessment with Gaussian processes}



\author[mymainaddress,mysecondaryaddress]{Morgane Menz}

\author[mysecondaryaddress]{Sylvain Dubreuil}
\author[mysecondaryaddress]{J\'{e}r\^{o}me Morio}
\author[mymainaddress]{Christian Gogu}

\author[mysecondaryaddress]{Nathalie Bartoli}
\author[mysecondaryaddress]{Marie Chiron}


\address[mymainaddress]{Universit\'{e} de Toulouse, UPS, CNRS, INSA, Mines Albi, ISAE, Institut Cl\'{e}ment Ader (ICA), 3 rue Caroline Aigle, 31400 Toulouse, France}
\address[mysecondaryaddress]{ONERA/DTIS, Universit\'{e} de Toulouse, F-31055 Toulouse, France}

\begin{abstract}

Running a reliability analysis on engineering problems involving complex numerical models can be computationally very expensive, requiring advanced simulation methods to reduce the overall numerical cost. Gaussian process based active learning methods for reliability analysis have emerged as a promising way for reducing this computational cost. The learning phase of these methods consists in building a Gaussian process surrogate model of the performance function and using the uncertainty structure of the Gaussian process to enrich iteratively this surrogate model. For that purpose a learning criterion has to be defined. Then, the estimation of the probability of failure is typically obtained by a classification of a population evaluated on the final surrogate model. Hence, the estimator of the probability of failure holds two different uncertainty sources related to the surrogate model approximation and to the sampling based integration technique. In this paper, we propose a methodology to quantify the sensitivity of the probability of failure estimator to both uncertainty sources. This analysis also enables to control the whole error associated to the failure probability estimate and thus provides an accuracy criterion on the estimation. Thus, an active learning approach integrating this analysis to reduce the main source of error and stopping when the global variability is sufficiently low is introduced. The approach is proposed for both a Monte Carlo based method as well as an importance sampling based method, seeking to improve the estimation of rare event probabilities. Performance of the proposed strategy is then assessed on several examples. 
\end{abstract}

\begin{keyword}
\texttt{Failure probability, Reliability, Monte Carlo, Importance Sampling, Active learning, Gaussian process, Sensitivity analysis, Classification}
\end{keyword}

\end{frontmatter}


\section{Introduction}
Engineering systems are subject to numerous uncertainties that imply a probability that these systems can fail. Reliability analyses seek to determine this probability of failure in order to understand, certify or improve their design.

Let us consider a system having input parameters $\bm{x}$ affected by uncertainties. A failure mode is characterized by a criterion defined by a performance function $G(\bm{x})$. 
By convention, a negative value of this function corresponds to the failure of the system, whereas a positive value means that the system is operational in this configuration. The limit between the failure and the safety domain is called the limit state.

Numerous reliability analysis techniques, i.e. techniques to estimate the probability of failure, can be found in the literature such as analytic approximations (FORM/SORM) \cite{Lemair_structreliability}, sampling methods based on Monte Carlo Simulations techniques \cite{melchers_structural_2018}, surrogate-based reliability analysis methods \cite{bucher_most_2008}, which can be adaptive or not. Adaptive approaches have been proposed in particular for Gaussian process surrogates \cite{vazquez_sequential_2009,echard_combined_2013,bichon_efficient_2008,bichon_efficient_2011,fauriat_ak-sys:_2014,li_bayesian_2012, bect_sequential_2012,dubourg_metamodel-based_2013}, support vector machines \cite{basudhar_reliability_2013, bourinet_assessing_2011} and polynomial-chaos-based Kriging \cite{schobi_r._rare_2017}. Considering any sampling technique, the probability of failure is obtained by a classification of the samples. The latter have to be evaluated first in order to be classified. This evaluation phase can be numerically very expensive for complex models. Gaussian process-based adaptive sampling methods for reliability analysis represent one of the promising ways for reducing this computational cost. 

Gaussian process-based adaptive sampling methods consist in building a Gaussian process surrogate model (or Kriging surrogate model) \cite{rasmussen_gaussian_2006,forrester_recent_2009} of the performance function and using the uncertainty structure of the Gaussian process to enrich iteratively this surrogate model. For that purpose, a learning criterion is used to select enrichment points at each iteration of the learning phase in order to better learn the limit state. Then, the estimation of the probability of failure is typically obtained by a classification of a set of Monte Carlo samples evaluated on the final surrogate model.  

Several adaptive methods have been proposed along these lines, such as the efficient global reliability analysis (EGRA) by Bichon et. al \cite{bichon_efficient_2008} or Active learning reliability method combining Kriging and Monte Carlo Simulations (AK-MCS) by Echard et. al \cite{echard_ak-mcs:_2011}. 
Other methods have also been presented to address specific problems such as small failure probabilities (rare events) estimations \cite{echard_combined_2013,huang_assessing_2016,lelievre_ak-mcsi:_2018,balesdent_kriging-based_2013, dubourg_metamodel-based_2013,li_bayesian_2012,cadini_bayesian_2016}, multiple failure regions problems \cite{cadini_improved_2014,lv_new_2015,zhang_reif:_2019,wang_reak:_2019}
or systems failure probabilities assessment \cite{fauriat_ak-sys:_2014,gaspar_adaptive_2017,bichon_efficient_2011,vazquez_sequential_2009,bect_sequential_2012,perrin_active_2016}.

In adaptive surrogate based methods, the estimator of the probability of failure is affected by two different uncertainty sources related to the surrogate model approximation and to the Monte Carlo (MC) based integration technique. A first important question is to know what is the share of these two sources of uncertainty in the final probability of failure estimate. It is notably important for the stopping criteria of the adaptive enrichment of the surrogate. Traditionally these stopping criteria are mainly defined to build a very accurate surrogate model around the limit state surface, such that only a maximal tolerated error on the Monte Carlo based integration can be imposed. This is not an optimal strategy however, since it may be over-conservative to seek an extremely accurate Gaussian process approximation in the vicinity of the limit state, if the main remaining source of error stems from the limited number of samples in the MC based integration.

Some investigations have already been carried out to take into account the Gaussian process accuracy on the quantity of interest (failure probability estimator), instead of the Gaussian process local error in the vicinity of the limit state. In \cite{schobi_r._rare_2017} some bounds of the estimator or in \citep{wang_esc:_2019} an approximation of the estimation relative error are used. None of these works have however thought to specifically separate the two previously described sources of error. 

Such an approach would allow to have access to the total variance of the estimator, taking account of the sampling and the Gaussian process model uncertainties, to validate the accuracy of the probability of failure estimation. Moreover, an alternative approach for the enrichment would be to seek to individually quantify the error induced by the surrogate model approximation and the error induced by the limited sampling set at each iteration and accordingly adjust between improvement of the surrogate model or enrichment of the sampling set.

In this paper, we propose to analyse both the Monte Carlo sampling and the surrogate model influence on the probability of failure estimator with variance based sensitivity indexes. We show that it is possible to estimate them numerically. It enables us to analyse quantitatively the source of uncertainty that has to be reduced to improve the accuracy of the failure probability estimate. We finally propose a new reliability assessment algorithm that integrates this analysis to focus on the main source of uncertainty during the learning phase and also provides a stopping criterion based on the whole error associated to the failure probability estimate.

The rest of the paper is organized as following. First, we present estimators of the sensitivity indices of the probability of failure and also of the total variance. Then, we propose a reliability analysis algorithm that integrates this sensitivity analysis to adaptively improve the major source of uncertainty during the learning phase. A stopping criterion based on the total variance estimation is also proposed for this algorithm. Finally, we present an extension of the method adapted to tackle rare events probability estimation problems.

\section{Reliability analysis}\label{sec:methods}

\subsection{General setting of reliability analyses}\label{sec_MCS}

Let $x_1,...,x_m$ be the $m$ uncertain parameters that are input to the reliability problem. These parameters are modeled by an absolutely continuous random vector $\bm{X}$ of random variables $X^k,$ $k=1,\ldots,m$ characterized by a joint probability distribution with probability density function $f_{\bm{X}}$.

In the context of reliability, the output of interest is the performance function $G:\mathbbm{R}^{m} \rightarrow \mathbbm{R}$. This function characterizes the failure of a system. Hence the domain of failure reads $\mathcal{D}_f=\{\bm{x} \in \mathbbm{R}^{m}, G(\bm{x}) \leq 0\}$, the domain of safety reads $\{\bm{x} \in \mathbbm{R}^{m}, G(\bm{x}) > 0\}$ and the limit state is $\{\bm{x}\in \mathbbm{R}^{m}, G(\bm{x}) = 0\}$. 
The failure probability $P_f$ is then defined as: 
\begin{equation}\label{eq:pf_def_int}
P_f= \mathbbm{E}_{f_{\bm{X}}}\left[\mathbbm{1}_{G(\bm{X})\leq 0} \right] = \int_{\mathbbm{R}^m} \mathbbm{1}_{G(\bm{x})\leq 0} f_{\bm{X}}(\bm{x}) \bm{dx} 
\end{equation}
where $\mathbbm{1}_{G(\bm{x})\leq 0}$ is an indicator function.
Several methods exist to evaluate this probability \cite{morio2014survey}. One of the simplest method is Monte Carlo Simulation (MCS). It consists in the generation of $n_{MC}$ random independent and identically distributed (i.i.d) samples $\bm{X}_1,...,\bm{X}_{n_{MC}}$ with distribution $f_{\bm{X}}$ and computing an estimation of the failure probability using these samples. As the failure probability can be expressed as a mathematical expectation (see Eq.~\eqref{eq:pf_def_int}), the law of large numbers suggests to build its estimator as the empirical mean of $\left(\mathbbm{1}_{G(\bm{X}_i)\leq 0}\right)_{i=1,\ldots,n_{MC}}$. 

An estimation $\hat{P}_f^{MC}$ of the failure probability $P_f$ is then given by: 
\begin{equation}
\hat{P}_f^{MC}=\frac{1}{n_{MC}}\sum_{i=1}^{n_{MC}}\mathbbm{1}_{G(\bm{x}) \leq 0}(\bm{X}_i)
\end{equation}
The variance of this estimator is given by: 
\begin{equation}
   Var \left(\hat{P}_f^{MC}\right)=\frac{Var\left(\mathbbm{1}_{G(\bm{X}_1)\leq 0}(\bm{X}_1)\right)}{n_{MC}}
  \label{eq:var_MC}
\end{equation}
In practice, the coefficient of variation (COV) can be used to quantify the uncertainty of the estimated failure probability. It can be estimated as following: 
\begin{equation}
\widehat{COV_{{P_f}^{MC} }}=\sqrt{\frac{(1-\hat{P}_f^{MC})}{n_{MC} \hat{P}_f^{MC}} }\label{eq:COV_mc}
\end{equation}
It can be seen in  Eq.~\eqref{eq:COV_mc} that for a failure probability of $10^{-n}$, $10^{n+2}$ simulations are needed to obtain an estimated coefficient of variation of about $10\%$.

Hence, MCS based classification methods need a lot of simulations to estimate small failure probabilities. In order to avoid the evaluation of a complex performance function $G(\bm{x})$ on a whole Monte Carlo population, an approximation by a surrogate model, denoted $\hat{G}(\bm{x})$, of this function can be used instead. However the accuracy of the surrogate model needs to be controlled in the regions near the limit state to perform a reliable classification of the MCS samples. For this purpose, Gaussian process regression based adaptive sampling methods allow to construct and enrich a Gaussian process by using the uncertainty structure of this type of surrogate to adaptively add learning points in regions that contribute significantly to the probability of failure estimate. 

\subsection{Reliability analysis using a Gaussian process}
Reliability analysis with a surrogate model relies mainly on four elements:
\begin{itemize}
    \item[$\bullet$] the type of surrogate model. Throughout the article, the surrogate model $\hat{G}(\bm{x})$ is assumed to be a Gaussian process and we will review its basics in Sec.~\ref{krig}.
    \item[$\bullet$]the sampling approach. In this article, we only consider Monte Carlo based sampling approaches such as MCS or importance sampling (see Sec.~\ref{sec_MCS} and Sec.~\ref{sec_IS}). 
    \item[$\bullet$]the surrogate model enrichment criterion used to most appropriately enrich the surrogate model in order to achieve an accurate approximation of the limit state.  
            \item[$\bullet$]the algorithm stopping criterion, that is set to determine when the  surrogate model learning is sufficient to obtain an accurate classification of the samples. 
\end{itemize}

In the introduction, many Gaussian process active learning methods are mentioned. Here, we are interested in methods that consider a population of candidate samples for the learning. This strategy has been first proposed in the AK-MCS \cite{echard_ak-mcs:_2011} method. Other methods have then been proposed in order to address more complex reliability problems as discussed in the introduction. In Sec.~\ref{sec:GP_rel}, some enrichment criteria and their corresponding stopping criteria used in these methods will be analyzed.

 \subsubsection{The Gaussian Process surrogate model}\label{krig}

Gaussian process regression, introduced in geostatistics by Krige \cite{krige_statistical_1951} and formalized later by Matheron \cite{matheron_principles_1963}, is a method of interpolation in which the interpolated function is modeled by a Gaussian process.

A Kriging or Gaussian process interpolation (GP) \cite{rasmussen_gaussian_2006}, denoted by $\mathcal{G}$, is fully characterized by its mean function $m(\bm{x})$ and a kernel (or covariance function) $k(\cdot,\cdot)$. Hence, the GP prior can be defined as: 
\begin{equation}
    \mathcal{G}(\bm{x})=m(\bm{x})+Z(\bm{x}) 
\end{equation}
where:
\begin{itemize}
    \item  $m(\bm{x})=\bm{f}(\bm{x})^T\bm{\beta}$ with $\bm{f}(\bm{x})$ a vector of basis functions and $\bm{\beta}$ the associated regression coefficients.
    
    \item $Z(\bm{x})$ a stationary zero mean Gaussian process with the variance $\sigma_Z^2$ such that the kernel defining the GP is $$k(\bm{x},\bm{x'})=\mbox{CoVar}(\mathcal{G}(\bm{x}),\mathcal{G}(\bm{x'}))=\sigma_Z^2 r_{\theta}(\bm{x},\bm{x'})$$ 
    $ r_{\theta}(\bm{x},\bm{x'})$ being a correlation function defined by the hyperparameter set $\bm{\theta}$ and $\mbox{CoVar}(\cdot,\cdot)$ being the covariance function between two points. In this paper, only stationary kernels are used that means kernels that are functions of $\bm{d}=|\bm{x}-\bm{x'}|$ (i.e. $ r_{\theta}(\bm{x},\bm{x'})= r_{\theta}(\bm{d})$).
\end{itemize}

Several kernel models exist to define the correlation function.
The squared exponential (SE) is probably the most widely-used kernel and has the form:
\begin{equation}
    r^{\text{SE}}(\bm{d})= \exp{\left[-\frac{\bm{d}^2}{\bm{l}^2}\right]} \label{eq:kern_SE}
\end{equation}
with parameter $\bm{l}$ the correlation length-scale, such that $\bm{l}^2=\nicefrac{1}{\bm{\theta}}$.  

This correlation function is infinitely differentiable, which means GPs constructed with this kernel are very smooth. Other kernels exist such as the ones of the well-known Mat\'{e}rn class. 
In this paper, the Mat\'{e}rn $\nicefrac{5}{2}$ will be used, whose expression is given by: 
\begin{equation}
    r^{\text{Mat52}}(\bm{d})=(1+\frac{\sqrt{5}\bm{d}}{\bm{l}}+\frac{5\bm{d}^2}{3\bm{l}^2} )\exp{\left[-\frac{\sqrt{5}\bm{d}}{\bm{l}}\right]} \label{eq:kern_mat52}
\end{equation}

Finally, the hyperparameters $\bm{\bm{\theta}}$, $\sigma_Z$ and $\bm{\beta}$ must be estimated to approximate the response for any unknown point of the domain. For a fixed kernel type, several techniques exist to obtain the optimal values of these hyperparameters, for example by Maximum Likelihood Estimation \cite{jones_taxonomy_2001} or cross-validation \cite{rasmussen_gaussian_2006}. 

 The prior distribution of $\mathcal{G}$ is considered to be Gaussian. Hence, the posterior distribution $\mathcal{G}_n$ of  $\mathcal{G}$ knowing the observations $\{\bm{x}_{doe}=(\bm{x}_1,...,\bm{x}_n),\bm{y}=G(\bm{x}_{doe})\}$ is Gaussian $\mathcal{G}_n=\mathcal{G}|(\bm{x}_{doe},y) \sim GP(\mu_{n}(\cdot), \sigma_{n}^2(\cdot,\cdot))$. 
 The GP predictor $\hat{G}(\bm{x})$ associated to the response has its mean value $\mu_{n}(\bm{x})$ and covariance $\sigma_{n}^2(\bm{x},\bm{x'})$ given by:
 \begin{equation}\label{eq:mean}
     \mu_{n}(\bm{x})=\bm{f}(\bm{x})^T\bm{\hat{\beta}}+\bm{k}(\bm{x})^T\bm{C}^{-1}(\bm{y}-\bm{F}\bm{\hat{\beta}})
 \end{equation}

  \begin{equation}\label{eq:variance}
     \sigma_{n}^2(\bm{x},\bm{x'})=\bm{k}(\bm{x},\bm{x'}) - \begin{pmatrix} \bm{k}(\bm{x})^T & \bm{f}(\bm{x})^T \end{pmatrix}  \begin{pmatrix} \bm{C} & \bm{F}^T \\ \bm{F} & \bm{0} \end{pmatrix}^{-1} \begin{pmatrix} \bm{k}(\bm{x'}) \\ \bm{f}(\bm{x'}) \end{pmatrix}  
 \end{equation}
 where $\bm{k}(\bm{x})=(k(\bm{x},\bm{x}_1),\ldots,k(\bm{x},\bm{x}_n))^T$, $\bm{F}$ is the matrix with row $i$ equals to $\bm{f}(\bm{x}_i)^T$, $\bm{C}:=(k(\bm{x}_i,\bm{x}_j))_{i,j}$ is the covariance matrix between the observations, and  $\bm{\hat{\beta}}=(\bm{F}^T\bm{C}^{-1}\bm{F})^{-1}\bm{F}^T\bm{C}^{-1}\bm{y}$.

 In the following section, the principle of adaptive sampling reliability analysis methods based on an active learning of a Gaussian process will be presented. 
 
\subsubsection{Gaussian process based reliability methods}\label{sec:GP_rel}

Gaussian process based reliability methods consist in the learning of a GP of the performance function. Therefore, the Gaussian process is iteratively enriched throughout a learning process in order to be very accurate in the vicinity of the limit state. The constructed surrogate model is thus well-suited for the classification of samples and allows to obtain an accurate estimation of the probability of failure.

The selection of the best point, with respect to the improvement of the GP approximation of the limit state, among all candidate samples, is based on a specific learning criterion. These learning criteria are built based on learning functions used to determine the most relevant point to evaluate the performance function at each iteration of the algorithm.
Many learning functions exist but we will focus here on two classic learning functions that are the functions $U$ and $EFF$.
The function $U$ proposed in \cite{echard_ak-mcs:_2011} is given by: 
\begin{equation}
    U(\bm{x})=\frac{|\mu_n(\bm{x})|}{\sigma_n(\bm{x})}
\end{equation}
The function $U$ quantifies the distance, expressed in GP standard deviations, between
the prediction mean and the estimated limit state. This criterion is evaluated on the Monte Carlo population and the performance function $G$ is computed on the sample $\bm{x}$ that minimizes $U$. Hence, this new observation is used to enrich the GP. In AK-MCS, the learning stopping condition for $U$ is defined as $\displaystyle{\min_{\bm{x}}(U(\bm{x}))}\geq 2$, corresponding to a probability $\Phi(-2)=0.023$ of making a mistake on the sign of the performance function value at $\bm{x}$.

Another learning criterion is the expected feasibility function $EFF(\bm{x})$, initially coming from the EGRA method \cite{bichon_efficient_2008}, and is given by the following expression: 
    \begin{align}
    \begin{split}
    EFF(\bm{x})&=\mu_n(\bm{x}) \left[2\Phi\left(-\frac{\mu_n(\bm{x})}{\sigma_{n}(\bm{x})}\right) - \Phi\left(-\frac{\epsilon +\mu_n(\bm{x})}{\sigma_{n}(\bm{x})}\right) -\Phi\left(\frac{\epsilon-\mu_n(\bm{x})}{\sigma_{n}(\bm{x})}\right) \right] \\
   & - \sigma_{n}(\bm{x})\left[2\phi\left(-\frac{\mu_n(\bm{x})}{\sigma_{n}(\bm{x})}\right) - \phi\left(-\frac{\epsilon +\mu_n(\bm{x})}{\sigma_{n}(\bm{x})}\right) -\phi\left(\frac{\epsilon-\mu_n(\bm{x})}{\sigma_{n}(\bm{x})}\right) \right] \\
   & +\epsilon \left[ \Phi\left(\frac{\epsilon -\mu_n(\bm{x})}{\sigma_{n}(\bm{x})}\right) -\Phi\left(-\frac{\epsilon+\mu_n(\bm{x})}{\sigma_{n}(\bm{x})}\right) \right]
   \end{split} \label{eq:EFF_func}
\end{align}

where $\Phi(\cdot)$ is the standard normal cumulative distribution function and $\phi(\cdot)$ the standard normal density function. The $EFF$ criterion provides an indication of how well the true value of the performance function at a point $\bm{x}$ is expected to satisfy the constraint $G(\bm{x})=0$ (i.e. expected to belong to the limit state). 
The parameter $\epsilon$ defines a region in the close vicinity of the threshold $\pm \epsilon$ over which the expectation is computed. In EGRA and AK-MCS+$EFF$, the expected feasibility function is built with $\epsilon=2\sigma_n$.

At each iteration, the next best point to evaluate is then the candidate sample whose $EFF$ value is maximum. The learning stopping condition is based on a stopping value of the learning criterion and is defined as:$$\max_{\bm{x}}(EFF(\bm{x}))\leq 0.001$$

For each method, a probability of failure is estimated once the learning phase is completed using the final conditioned Gaussian process $\mathcal{G}_n$. The probability of failure estimation on a Monte Carlo population of $n_{MC}$ samples  $\tilde{\bm{X}}=(\bm{X}_i)_{i=1,\cdots,n_{MC}}$ with $\bm{X}_i$ i.i.d. with the same probability distribution as $\bm{X}$ is then given by: 

\begin{equation}\label{eq:pf_mc_gp_mean}
\hat{P}_f^{MC}(\tilde{\bm{X}}) =\frac{1}{n_{MC}} \sum_{i=1}^{n_{MC}} \mathbbm{1}_{\mu_n(\bm{X}_i)\leq 0} (\bm{X}_i) 
\end{equation}

Moreover, as the learning is done on a population of candidate samples, once the learning phase is completed, the algorithm may generate additional samples in order to satisfy a maximum COV of the probability of failure estimation. Then, the learning phase is restarted in order to have a GP suited for the whole Monte Carlo population.

Here, the part of failure probability variance due to the GP is neglected as the learning criteria are conceived to build a very confident GP model in terms of classification accuracy, which justifies the use of the mean $\mu_n(\bm{X}_i)$ of the GP predictor in the estimator expression given by Eq.~\eqref{eq:pf_mc_gp_mean}. In fact, learning stopping conditions are in general very conservative, which probably leads to an overquality of the GP when compared to the sampling variance.

However, the probability of failure estimator is actually a random variable and its variance depends on both the uncertainty of the integration by MCS and the uncertainty of the performance function surrogate model approximation. 

Some investigations to take into account the GP accuracy on a quantity of interest have been carried out. For example, Le Gratiet proposed in \cite{le_gratiet_bayesian_2014} to provide confidence intervals of Sobol indices estimated by GP regression and Monte Carlo integration. Therefore, a quantification of the contribution of both uncertainty sources to the Sobol indices estimators variability is proposed in \cite{le_gratiet_bayesian_2014}.
In \cite{zhu_reliability_2016,el_haj_improved_2021}, a learning function is proposed that is  based on the contribution of a point of the MC population, considering the dependencies to other samples, to the error of the failure probability estimation.  
In \cite{schobi_r._rare_2017}, Sch\"{o}bi proposed to use bounds of the probability of failure estimator $\hat{P}_f$ in an active learning algorithm for reliability analysis to define a learning stopping condition. In Sch\"{o}bi's work, the bounds were computed by classifying the points of the population using their prediction bound values.
In the next section, we provide new measures of the influence on the probability of failure of the use of numerical integration by MCS and surrogate model approximations based on a variance decomposition.

\section{Measure of failure probability sensitivity to GP and MC estimation uncertainties}

In the previous section, we note that the estimation of the probability of failure depends on both the Monte Carlo population and the GP approximation of the performance function. The idea is to consider the Monte Carlo estimation of Eq.~\eqref{eq:pf_mc_gp_mean} but to replace the GP mean $\mu_n$ by the GP $\mathcal{G}_n$. 

\begin{equation}\label{eq:pf_gp_mc2}
\hat{P}_f(\tilde{\bm{X}}, \mathcal{G}_n) =\frac{1}{n_{MC}} \sum_{i=1}^{n_{MC}} \mathbbm{1}_{\mathcal{G}_n(\bm{X}_i)\leq 0} (\bm{X}_i) = \Gamma(\tilde{\bm{X}},\mathcal{G}_n)
\end{equation}
where $\Gamma$ is in fact a deterministic scalar function which has two random inputs: the Monte Carlo sampling $\tilde{\bm{X}}$ and the conditioned GP $\mathcal{G}_n$. 

In order to assess the contributions of each of both uncertainty sources $\tilde{\bm{X}}$ and $\mathcal{G}_n$ on the variability (i.e. variance) of $\hat{P}_f$ separately, we can refer to the variance decomposition expression \cite{iooss_review_2015} which is a classical tool in sensitivity analysis:  
\begin{equation}
Var_{\mathcal{G}_n,\tilde{\bm{X}}}\left(\hat{P}_f(\tilde{\bm{X}},\mathcal{G}_n) \right) =V_{\tilde{\bm{X}}} +V_{\mathcal{G}_n} + V_{\mathcal{G}_n,\tilde{\bm{X}}}
\label{eq:Var_decomp}
\end{equation}
where: 
\begin{itemize}
\item $V_{\tilde{\bm{X}}}=Var_{\tilde{\bm{X}}}\left(\mathbb{E}_{\mathcal{G}_n}\left[\hat{P}_f|\tilde{\bm{X}}\right] \right)$ measures the influence of the Monte Carlo sampling on the variance of $\hat{P}_f$,
\item $V_{\mathcal{G}_n}=Var_{\mathcal{G}_n}\left(\mathbb{E}_{\tilde{\bm{X}}}\left[\hat{P}_f|\mathcal{G}_n\right] \right)$ measures the influence of the GP uncertainty on the variance of $\hat{P}_f$,
\item $V_{\mathcal{G}_n,\tilde{\bm{X}}}=Var_{\mathcal{G}_n,\tilde{\bm{X}}}\left(\mathbb{E} \left[\hat{P}_f|\mathcal{G}_n,\tilde{\bm{X}} \right] \right) - V_{\mathcal{G}_n}-V_{\tilde{\bm{X}}}$ measures the joint contribution of both Monte Carlo and GP uncertainties on the variance of $\hat{P}_f$.
\end{itemize}

\subsection{Variance contributions estimation} \label{sec:var_estimators}
As the GP enrichment points are chosen among the MC population, $\tilde{\bm{X}}$ and $\mathcal{G}_n$ are theoretically not independent. However, as the samples of the MC population used to learn the GP only represent a small part of the population. The estimators developed in the next sections are based on an independence hypothesis.  

\subsubsection{Variance estimator}\label{sec:theorie_var_int}

Let us assume we have a random i.i.d sample of size $n_s$ $(Z_1,\ldots,Z_{n_s})$  of a random variable $Z$ following an unknown distribution. 
The mean of $Z$ is approached by the empirical mean over the $n_s$ samples denoted by $\overline{Z_{n_s}}$. The empirical variance of a random vector $Z$, denoted $\widehat{Var}(Z)$ throughout the paper, is defined by: 
\begin{equation}
     \widehat{Var}(Z)=\frac{1}{n_s-1}\sum_{i=1}^{n_s}\left(Z_i- \overline{Z_{n_s}}\right)^2
\end{equation}
However, this estimator is actually a random variable due to sampling variation. Therefore, it is more convenient to compute confidence interval estimations that contain with a given chosen probability the real value of the variance. 
 The estimated asymptotic confidence interval  $\left[ \widehat{Var^{inf}}(Z),\widehat{Var^{sup}}(Z) \right]$ of the variance of level $1-\alpha$ is given according to the central limit theorem by:

   \begin{equation}
\left[\widehat{Var}(Z)- k\frac{\sqrt{n_s\widehat{Var}\left( \left(Z_i- \overline{Z_{n_s}}\right)^2 \right)}}{n_s-1} ;\quad \widehat{Var}(Z) + k \frac{\sqrt{n_s\widehat{Var}\left( \left(Z_i- \overline{Z_{n_s}} \right)^2 \right) }}{n_s-1} \right] \label{eq:variance_bounds_theorie}
\end{equation}
 where $k$ is the quantile of order $1-\alpha$ of the reduced centred normal distribution.
 
Hence, if we want to identify the most influential source of uncertainty this property of the variance estimator allows us to compare the bounds of the confidence intervals of $V_{\tilde{\bm{X}}}$ and $V_{\mathcal{G}_n}$. Moreover it allows to evaluate the exact number of simulations required to have disjoint confidence intervals.

\subsubsection{Expression of $V_{\tilde{\bm{X}}}$ estimator}

First let us recall that the random variable $Y= \mathbbm{1}_{\mathcal{G}_n(\bm{x})\leq 0}(\bm{x})$ is a Bernoulli random variable $\mathcal{B}(p(\bm{x}))$ with parameter $p(\bm{x})=\mathbbm{P}\left[\mathcal{G}_n(\bm{x})\leq 0\right]=\Phi\left(-\frac{\mu_n(\bm{x})}{\sigma_n(\bm{x})}\right)$, the probability that $\bm{x}$ belongs to the failure domain according to the Gaussian process $\mathcal{G}_n$. 

The expected value of $\hat{P}_f$, given by Eq.~\eqref{eq:pf_gp_mc2}, knowing a Monte Carlo population of $n_{MC}$ samples $\tilde{\bm{X}}=(\bm{X}_i)_{i=1,..,n_{MC}}$ can be expressed as follows: 

\begin{align}
\begin{split}
  \mathbb{E}_{\mathcal{G}_n}\left[\hat{P}_f|\tilde{\bm{X}}\right] & =  \mathbbm{E}_{\mathcal{G}_n}\left[ \frac{1}{n_{MC}}\sum_{i=1}^{n_{MC}} \mathbbm{1}_{\mathcal{G}_n(\bm{X}_i)\leq 0}(\bm{X}_i)|(\bm{X}_i)_{i=1,..,n_{MC}}\right] \\
   & = \mathbbm{E}_{\mathcal{G}_n}\left[\frac{1}{n_{MC}}\sum_{i=1}^{n_{MC}} \mathcal{B}(p(\bm{X}_i))\right] \\
      & = \frac{1}{n_{MC}}\sum_{i=1}^{n_{MC}} \mathbbm{E}_{\mathcal{G}_n}\left[\mathcal{B}(p(\bm{X}_i))\right] \\
   & = \frac{1}{n_{MC}}\sum_{i=1}^{n_{MC}} p(\bm{X}_i)  \label{eq:E_G_X}
   \end{split}
\end{align}

Using the analytical expression of $\mathbb{E}_{\mathcal{G}_n}\left[\hat{P}_f|\tilde{\bm{X}}\right]$ given by Eq.~\eqref{eq:E_G_X}, the variance $V_{\tilde{\bm{X}}}$ can then be obtained by simulating: 
\begin{equation}
     V_{\tilde{\bm{X}}}=Var_{\tilde{\bm{X}}}\left(\frac{1}{n_{MC}}\sum_{i=1}^{n_{MC}} p(\bm{X}_i) \right)=\frac{Var_{\tilde{\bm{X}}}( p(\bm{X}))}{n_{MC}}  \label{eq:V_X} 
\end{equation}
The last equality is obtained as $p(\bm{X})$ is a continuous random variable between $0$ and $1$ and $p(\bm{X}_i)$ are i.i.d replications of it.

In practice, Eq.~\eqref{eq:V_X} is estimated on the Monte Carlo population used for the estimation of the probability of failure. Hence the estimator of $V_{\tilde{\bm{X}}}$, denoted by $\widehat{V_{\tilde{\bm{X}}}}$, on a MC population realization is given by:  
\begin{equation}
    \widehat{V_{\tilde{\bm{X}}}}=\frac{\widehat{Var_{\tilde{\bm{X}}}}( \bm{p}(\tilde{\bm{X}}))}{n_{MC}}=  \frac{1}{n_{MC}(n_{MC}-1)}\sum_{i=1}^{n_{MC}}\left(p(\bm{X}_i)-\frac{1}{n_{MC}} \sum_{j=1}^{n_{MC}} p(\bm{X}_j)\right)^2 \label{eq:V_X_est}
\end{equation}
and its $1-\alpha$ confidence interval estimated bounds can be expressed using the Eq.~\eqref{eq:variance_bounds_theorie} defined in Sec.~\ref{sec:theorie_var_int} and are given by:
\begin{equation}
\begin{array}{rcr}
      \widehat{V_{\tilde{\bm{X}}}^{inf}} &= & \frac{\widehat{Var_{\tilde{\bm{X}}}^{inf}}(\bm{p}(\tilde{\bm{x}}))}{n_{MC}}\\
      \widehat{V_{\tilde{\bm{X}}}^{sup}}&  =& \frac{\widehat{Var_{\tilde{\bm{X}}}^{sup}}(\bm{p}(\tilde{\bm{x}}))}{n_{MC}}
\end{array}    
\end{equation}

\subsubsection{Expression of $V_{\mathcal{G}_n}$ estimator}

The computation of the expected value of $\hat{P}_f$ knowing a realization of $\mathcal{G}_n$ can be interpreted as a classical Monte Carlo simulation for a deterministic model. Hence it follows this equality: 

\begin{align}
  \mathbb{E}_{\tilde{\bm{X}}}\left[\hat{P}_f|\mathcal{G}_n \right] & = 
   \mathbb{E}_{\tilde{\bm{X}}}\left[\frac{1}{n_{MC}}\sum_{i=1}^{n_{MC}} \mathbbm{1}_{\mathcal{G}_n(\bm{X}_i) \leq 0}(\bm{X}_i)|\mathcal{G}_n\right]\\
   & = P_f(\mathcal{G}_n)
\end{align}
with $P_f(\mathcal{G}_n)$ the probability of failure for a realization of $\mathcal{G}_n$. Classically, this probability of failure is approached by a Monte Carlo estimator  $\hat{P}_f^{MC}(\mathcal{G}_n) = \frac{1}{n_{MC}}\sum_{i=1}^{n_{MC}} \mathbbm{1}_{\mathcal{G}_n(\bm{X}_i) \leq 0}(\bm{X}_i)$, with $(\bm{X}_i)_{i=1,\ldots,n_{MC}}$ a Monte Carlo population realization. Hence, $V_{\mathcal{G}_n}$ can be numerically estimated by simulating different trajectories of $\mathcal{G}_n$ and computing the Monte Carlo estimator of $P_f$ for each simulated trajectory. 

Let $(G_i)_{1\le i\le n_t}$ be $n_t$ realizations of $\mathcal{G}_n$, then the $V_{\mathcal{G}_n}$ estimate is the empirical variance of the sample $\hat{P}_f^{MC}(G_i)_{1\le i\le n_t}$: 
\begin{equation}\label{eq:V_Gn_est}
     \widehat{V_{\mathcal{G}_n}}=\widehat{{Var}_{\mathcal{G}_n}}(\hat{P}_f^{MC}(\mathcal{G}_n))=\frac{1}{n_t-1} \sum_{i=1}^{n_t} \left(  \hat{P}_f^{MC}(G_i)  - \frac{1}{n_t} \sum_{i=1}^{n_t} \hat{P}_f^{MC}(G_i) \right)^2 
\end{equation}

Moreover its $1-\alpha$ confidence interval estimated bounds, also expressed using the Eq.~\eqref{eq:variance_bounds_theorie} defined in Sec.~\ref{sec:theorie_var_int}, are given by: 
\begin{equation}
\begin{array}{rcr}
        \widehat{V_{\mathcal{G}_n}^{inf}}&  =& \widehat{Var_{\mathcal{G}_n}^{inf}}(\hat{P}_f^{MC}(\mathcal{G}_n))\\
       \widehat{V_{\mathcal{G}_n}^{sup}}&  =& \widehat{Var_{\mathcal{G}_n}^{sup}}(\hat{P}_f^{MC}(\mathcal{G}_n))
\end{array}    
\end{equation}

In practice, the computation of conditioned GP realizations is prone to numerical issues. In \ref{appendix:gp_traj}, these numerical issues and their sources are exposed and an alternative method, presented in \cite{le_gratiet_bayesian_2014,villemonteix_informational_2009}, based on the simulation of an unconditioned Gaussian process is detailed.

\subsubsection{Expression of the total variance estimator}

Furthermore, the total variance of $\hat{P}_f$ can be estimated with: 
\begin{equation}
     \widehat{V_{tot}}= \widehat{Var_{\mathcal{G}_n,\tilde{\bm{X}}}} \left(\hat{P}_f(\tilde{\bm{X}},\mathcal{G}_n) \right)=\frac{1}{n_t-1} \sum_{i=1}^{n_{t}}\left(  \hat{P}_f(G_i,\tilde{\bm{X}_i})  - \frac{1}{n_{t}} \sum_{j=1}^{n_{t}} \hat{P}_f(G_j,\tilde{\bm{X}_j})\right)^2
\label{eq:v_tot_est}
\end{equation}

where $(G_i, \tilde{\bm{X}}_i), \; i=1,\ldots,n_{t}$ are $n_{t}$ realizations of $\mathcal{G}_n$ and Monte Carlo population $\tilde{\bm{X}}$ and $\hat{P}_f(G_i, \tilde{\bm{X}}_i)$ is the probability of failure estimation for the $i^{th}$ realization $(G_i, \tilde{\bm{X}}_i)$ of $\mathcal{G}_n$ and $\tilde{\bm{X}}$. $\widehat{V_{tot}}$ is the empirical variance of the sample $(\hat{P}_f(G_i, \tilde{\bm{X}}_i))_{1\le i\le n_t}$.

Moreover the $1-\alpha$ confidence interval estimated bounds of $\widehat{V_{tot}}$, also expressed using the operators defined in Sec.~\ref{sec:theorie_var_int}, are given by: 
\begin{equation}
\begin{array}{rcr}
        \widehat{V_{tot}^{inf}}&  =& \widehat{Var_{\mathcal{G}_n,\tilde{\bm{X}}}^{inf}}(\hat{P}_f(\mathcal{G}_n,\tilde{\bm{X}}))\\
       \widehat{V_{tot}^{sup}}&  =& \widehat{Var_{\mathcal{G}_n,\tilde{\bm{X}}}^{sup}}(\hat{P}_f(\mathcal{G}_n,\tilde{\bm{X}}))
\end{array}    
\end{equation}

In practice, the numerical cost of the $n_t$ estimations of $\hat{P}_f(G_i,\tilde{\bm{X}}_i) = \hat{P}_f(G_i(\tilde{\bm{X}}_i))$ can be quite high to get a sufficiently low variance of the estimator $\widehat{V_{tot}}$. Hence, we propose to use a bootstrap procedure \cite{archer1997sensitivity} to simulate several MC populations $\tilde{\bm{X}}_i$ from the population $\tilde{\bm{X}}$ on which the $n_t$ GP trajectories are computed.

Finally, the probability of failure is estimated by the mean over the $(\hat{P}_f(G_i, \tilde{\bm{X}}_i))_{1\le i\le n_t}$, i.e. by: 

\begin{equation}
    \hat{P}_f^t = \frac{1}{n_{t}} \sum_{i=1}^{n_{t}} \hat{P}_f(G_i,\tilde{\bm{X}_i})  \label{eq:pf_final0}
\end{equation}

Hence, estimated total COV, denoted by $COV_{tot}$, of an estimation $\hat{P}_f^{MC}$ of the probability of failure $\hat{P}_f$ obtained on a MC realization is thus given by: 
\begin{equation}
     \widehat{COV_{tot}}=\frac{\sqrt{\widehat{V_{tot}}}}{\hat{P}_f^t}
     \label{eq:cov_tot_est}
\end{equation}

Its $1-\alpha$ confidence interval estimated bounds are then estimated by: 
\begin{equation}
\begin{array}{rcr}
         \widehat{COV}_{tot}^{inf}&  =& \frac{\sqrt{ \widehat{V_{tot}^{inf}}}}{\hat{P}_f^t} \\[1.3ex]
        \widehat{COV}_{tot}^{sup}&  =& \frac{\sqrt{ \widehat{V_{tot}^{sup}}}}{\hat{P}_f^t}
        \label{eq:cov_int}
\end{array}    
\end{equation}

Finally notice that the joint contribution of the MC integration and the GP approximation uncertainties $V_{\mathcal{G}_n,\tilde{\bm{X}}}$ can then be estimated by computing the three previous estimators and applying the relation given by Eq.~\eqref{eq:Var_decomp}. By carrying out this calculation the independence hypothesis introduced at the beginning of Sec.~\ref{sec:var_estimators} can be empirically verified.

\subsection{ Motivation for developing a new approach illustrated on a benchmark case}  \label{sec:appli_4branches_1}
The idea here is to explain our motivation to propose a new approach in order to perform a trade-off between improving the GP and adding points to the MCS. To illustrate this point, a well known benchmark example is chosen.

 The application deals with the example of a series system with four branches, also studied in \cite{echard_ak-mcs:_2011,schueremans_benefit_2005}. The random variables $X_1$ and $X_2$ follow standard normal distributions. The performance function is given by: 

\begin{equation}
    y=\min_{x_1,x_2}
\left\{
\begin{array}{l l }
& 3+0.1(x_1+x_2)^2-\frac{(x_1+x_2)}{\sqrt(2)};\\
&3+0.1(x_1+x_2)^2+\frac{(x_1+x_2)}{\sqrt(2)};\\
& (x_1-x_2)+\frac{6}{\sqrt{2}}; \\
& (x_2-x_1)+\frac{6}{\sqrt{2}}
\end{array}
\right\}
\label{eq:4b_limitstate}
\end{equation}

\begin{figure}[h!]
    \centering
    \includegraphics[width=0.99\textwidth]{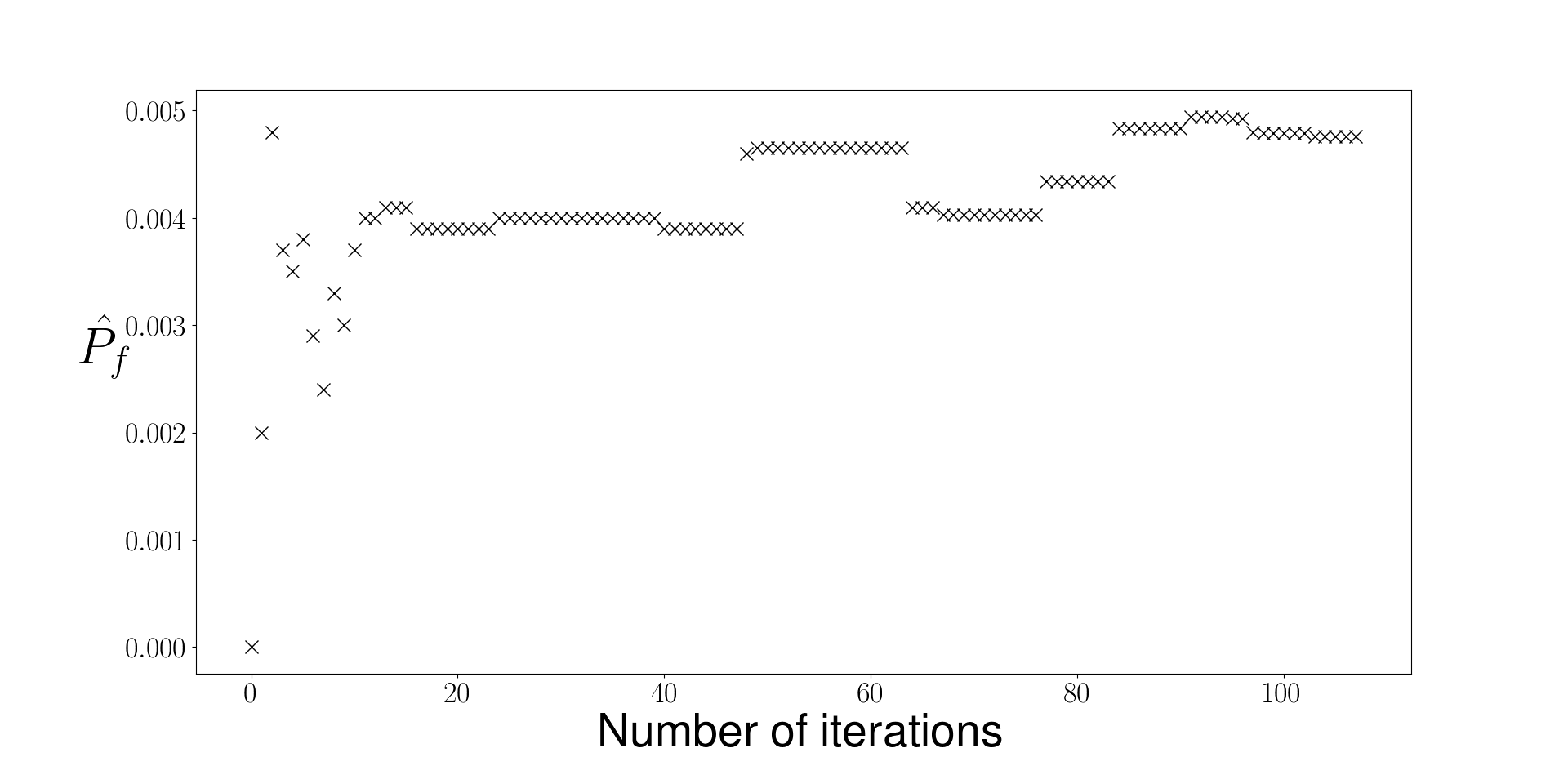}
    \caption{Evolution of the probability of failure estimation as a function of the number of iterations throughout a run of the algorithm AK-MCS+$EFF$ ($\hat{P}_f^{MC}=4.46 \times 10 ^{-3}(1.6\%)$) on the four branches test case} 
    \label{fig:akmcs_4b_pf}
\end{figure}

\begin{figure}[h!]
    \centering
    \includegraphics[width=0.99\textwidth]{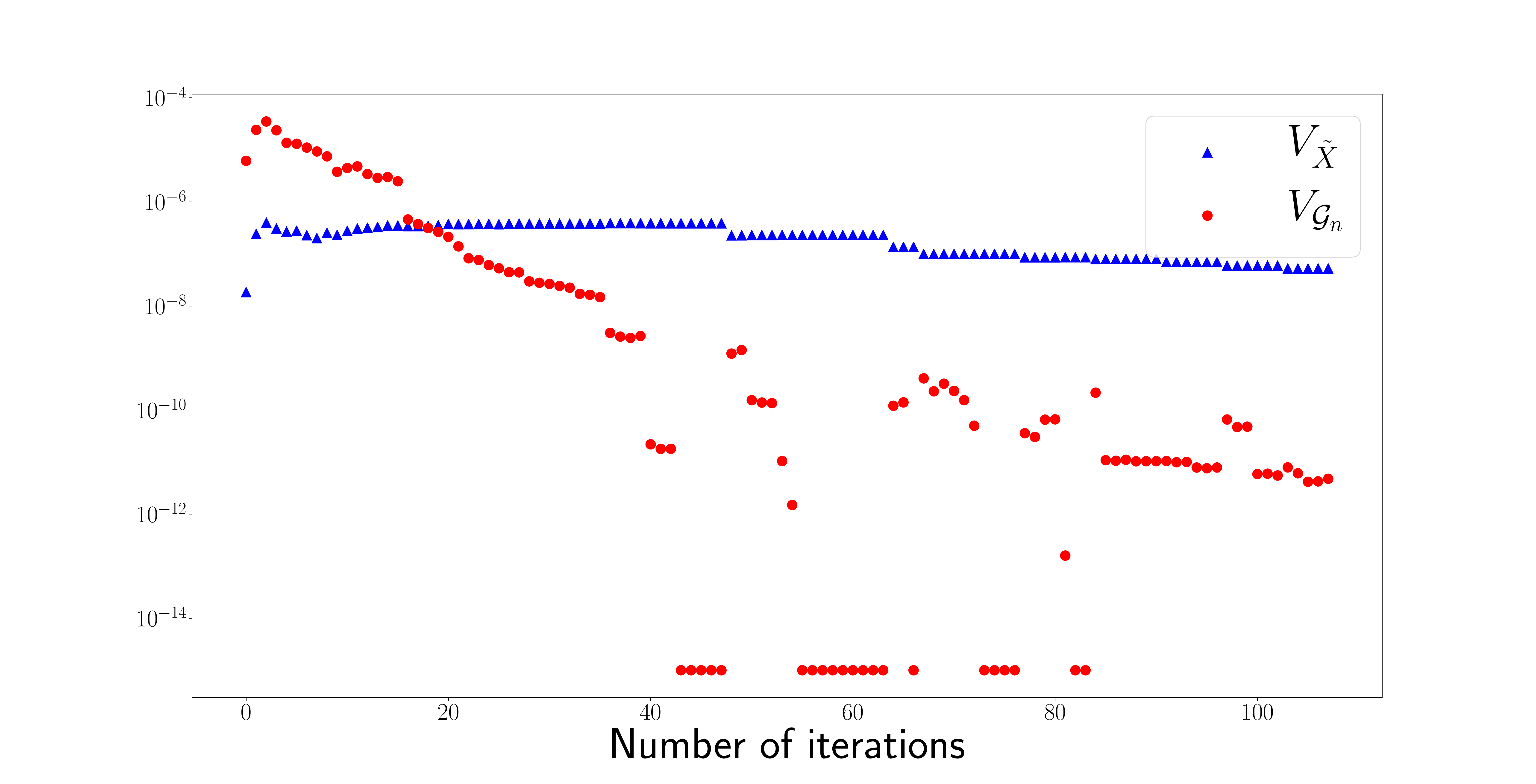}
    \caption{Evolution of the variances $V_{\tilde{\bm{X}}}$ and $V_{\mathcal{G}_n}$ estimations as a function of the number of iterations throughout a run of the algorithm AK-MCS+$EFF$ on the four branches test case}
    \label{fig:akmcs_4b_var}
\end{figure}

A run of the AK-MCS+$EFF$ algorithm gives an estimation of $P_f$ and the  corresponding MC coefficient of variation.
On this example, this algorithm was run with an initial Monte Carlo population of size $10^4$ and a maximum allowed coefficient of variation of $5\%$.
At each iteration of the algorithm, the variability due to the GP $\mathcal{G}_n$ and the Monte Carlo based integration $\tilde{\bm{X}}$ was estimated using respectively Eq.~\eqref{eq:V_Gn_est} and Eq.~\eqref{eq:V_X_est}. The evolution of the probability of failure and the corresponding variances estimations throughout the algorithm are respectively illustrated on Fig.~\ref{fig:akmcs_4b_pf} and Fig.~\ref{fig:akmcs_4b_var}.

In particular, it can be seen on Fig.~\ref{fig:akmcs_4b_var} that for the 18 first iterations the main contributor to the failure probability variance is the GP $\mathcal{G}_n$. Then, until the end of the run the principal source of variability is the Monte Carlo integration. However, the GP is still enriched after the $18^{th}$ iteration of this algorithm reaching 99 enrichment points when the algorithm converges. Moreover, it can be seen on Fig.~\ref{fig:akmcs_4b_pf} that the spread of $\hat{P}_f$ estimation values stops around the $15^{th}$ iteration and progresses then in the vicinity of the real value of the probability of failure. At the end of the run, the part of variability on $\hat{P}_f$ due to the  Monte Carlo integration is $1.1\times 10^4$ higher than the one attributed to the GP approximation. 

The variance comparison leads us to conclude that there is no need to learn the GP in a so accurate way and by avoiding this we can hope to save some unnecessary simulations of the performance function.

Hence, it could be interesting to integrate these measures of variance in the learning procedure to overcome the over-conservative learning of GP. The proposed method is detailed in the next section.

\section{Proposed method}\label{sec:new_method}
\renewcommand{\labelenumii}{\theenumii}
\renewcommand{\theenumii}{\theenumi.\arabic{enumii}.}

\subsection{General concept}
The new method consists in using the variance estimations obtained previously in the learning phase as decision criteria. 
 On the one hand, the contributions attributed to the Monte Carlo estimation and the GP to the variability of $\hat{P}_f$ can be used to decide whether to improve the GP or to increase the size of the sampling population. These contributions can be quantified using the variances estimators given by Eq.~\eqref{eq:V_X_est} for the MC integration contribution and by Eq.~\eqref{eq:V_Gn_est} for the GP approximation. 

On the other hand, the total variance on $\hat{P}_f$, whose estimator is given by Eq.~\eqref{eq:v_tot_est}, can be used as a criterion to stop the learning phase.

\subsection{Proposed algorithm }
\begin{figure}[h!]
    \centering
    \includegraphics[width=\linewidth]{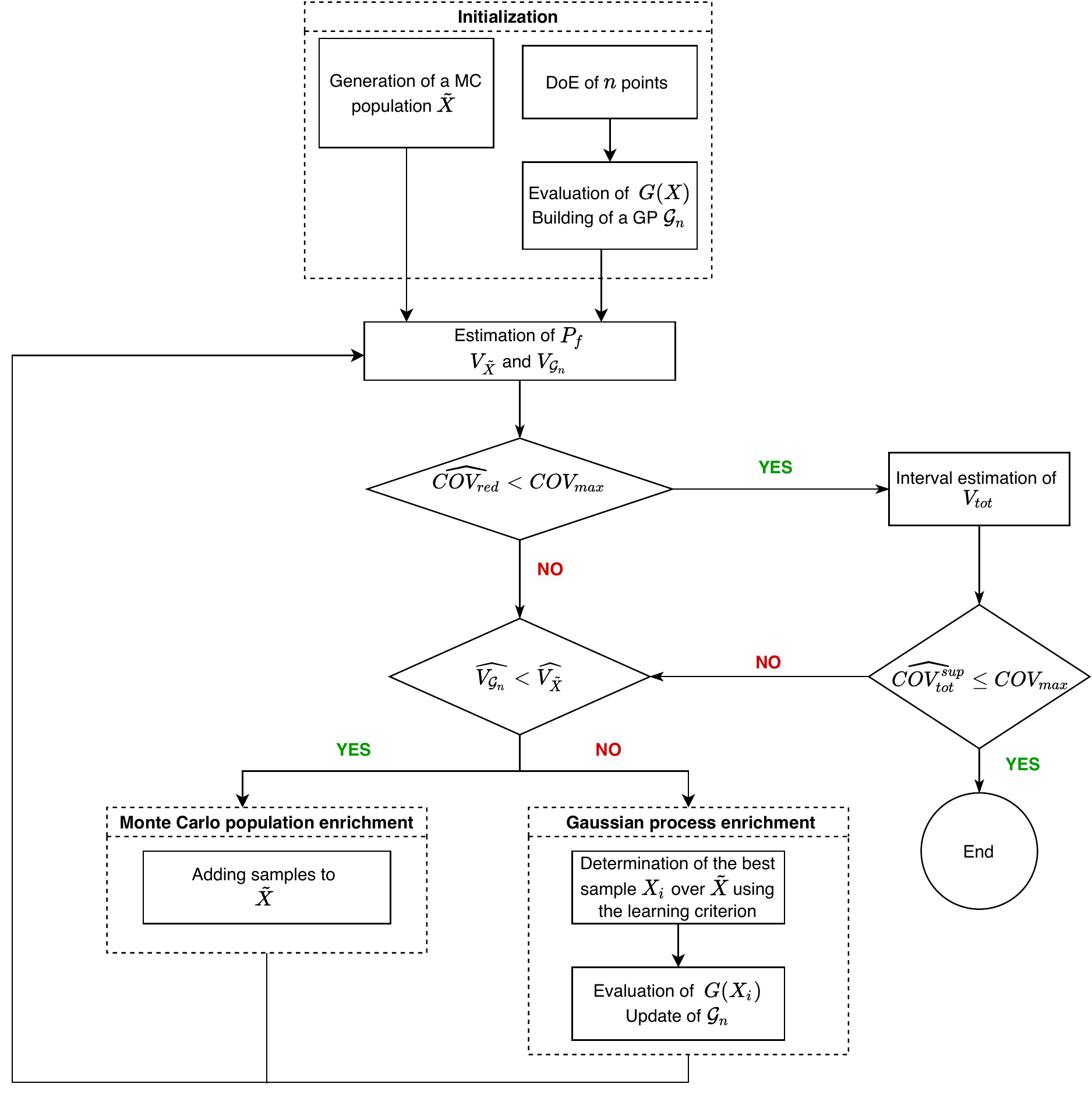}
    \caption{Flowchart of the learning algorithm Vb-AGP + MCS }
    \label{fig:sensitivity_mc}
\end{figure}

The proposed Variance based Active GP (Vb-AGP) learning procedure is summarized in Fig.~\ref{fig:sensitivity_mc} and the different stages are described below: 
\begin{enumerate}
    \item Generation of an initial Monte Carlo population $\tilde{\bm{x}}$ of $n_{MC}$ samples.
     \item Initial Design of Experiments (DoE) of $n$ samples defined using sampling methods such as Latin Hypercube Sampling (LHS). The performance function $G(\bm{x})$ is then evaluated at the $n$ samples. 
     \item Construction of a GP metamodel $\mathcal{G}_n(\bm{x})$ of the performance function $G(\bm{x})$ on the DoE.
     \item{Estimation of the failure probability $P_f$ on the Monte Carlo population $\tilde{\bm{x}}$ according to the following equation:
     \begin{equation}
\hat{P}_f^{MC}(\tilde{\bm{x}}) =\frac{1}{n_{MC}} \sum_{i=1}^{n_{MC}} \mathbbm{1}_{\mu_n(\bm{x}_i)\leq 0}
\end{equation}
    }
    
    \item Interval estimation of variances $V_{\tilde{\bm{X}}}$ and $V_{\mathcal{G}_n}$ using respectively Eq.~\eqref{eq:V_X_est} and Eq.~\eqref{eq:V_Gn_est}.
    
    Note that we seek to obtain
    $\left[\widehat{V_{\tilde{\bm{X}}}^{inf}},\widehat{V_{\tilde{\bm{X}}}^{sup}}\right] \cap \left[ \widehat{V_{\mathcal{G}_n}^{inf}},\widehat{V_{\mathcal{G}_n}^{sup}} \right] = \emptyset$, using the estimators given in  Sec.~\ref{sec:var_estimators}, in order to compare both variances values. Therefore new GP trajectories have to be simulated until the estimation of $V_{\mathcal{G}_n}$ confidence interval is sufficiently narrow.
    \item Compute $\widehat{COV_{red}}=\frac{\sqrt{\widehat{V_{\mathcal{G}_n}^{sup}}+\widehat{V_{\tilde{\bm{X}}}^{sup}}}}{\hat{P}_f^{MC}}$. 
    
    If $\widehat{COV_{red}}< COV_{max}$, with $COV_{max}$ a user defined maximum allowed total coefficient of variation, the algorithm goes to step 7 to verify that the total COV is below the maximum allowed value.
    
    Otherwise, the algorithm goes to step 8 in order to reduce the main source of uncertainty.
    
    \item Interval estimation of the total coefficient of variation $COV_{tot}$ using Eq.~\eqref{eq:cov_int}: increasing number of simulations until $COV_{max} \notin \left[  \widehat{COV}_{tot}^{inf}, \widehat{COV}_{tot}^{sup} \right]$. If $COV_{tot}^{sup} \leq  COV_{max}$ then the estimation $ \hat{P}_f^t$ of $\hat{P}_f$, whose expression is recalled in Eq.~\eqref{eq:pf_final}, is considered sufficiently accurate and the algorithm is stopped. 
    \begin{equation}
    \hat{P}_f^t = \frac{1}{n_{t}} \sum_{i=1}^{n_{t}} \hat{P}_f(G_i,\tilde{\bm{X}_i}) \label{eq:pf_final}
\end{equation}
where $(G_i, \tilde{\bm{X}}_i), \; i=1,\ldots,n_{t}$ are $n_{t}$ realizations of $\mathcal{G}_n$ and Monte Carlo population $\tilde{\bm{X}}$ used to compute the total COV estimation $COV_{tot}$.

Otherwise, the algorithm goes to next step. 
    \item If $\widehat{V_{\mathcal{G}_n}}  < \widehat{V_{\tilde{\bm{X}}}}$, new samples are added to the Monte Carlo population and the method goes back to step 4 to update the estimation of $P_f$.  
    
    Else if $\widehat{V_{\mathcal{G}_n}}  > \widehat{V_{\tilde{\bm{X}}}}$, the algorithm goes to step 9.
    \item The learning function $EFF(\bm{x})$ given by Eq.~\eqref{eq:EFF_func} is evaluated on the whole MC population to find the best candidate $\bm{x^*}$ to evaluate for enriching the GP metamodel. The performance function is computed on the sample $\bm{x^*}$ and the DoE is enriched with this new point $\bm{x^*}$. 
    Then the method goes to step 3 to update the GP model.
\end{enumerate}

The first stage of the stopping condition on learning consists in verifying the following equation:
        \begin{equation}
      \widehat{COV_{red}}=\frac{\sqrt{\widehat{V_{\mathcal{G}_n}^{sup}}+\widehat{V_{X}^{sup}}}}{\hat{P}_f^{MC}} < COV_{max} \label{eq:learning_crit_1}
    \end{equation}
    where $COV_{max}$ is a user defined maximum allowed total coefficient of variation,

The condition given by Eq.~\eqref{eq:learning_crit_1} corresponds to a condition on the approximation of the total variance of the $\hat{P}_f$, under the independence assumption. Indeed, throughout the learning, the joint contribution of both variables is never computed, since it would significantly increase the computational cost. If Eq.~\eqref{eq:learning_crit_1} is verified, the total variance $V_{tot}$ including the joint contribution can be estimated (i.e. in step 7 of the algorithm) to make sure that it respects the maximum variance allowed.

Let us now make a few comments about the choices made for this algorithm. 
First we have found that the learning function $EFF(\bm{x})$ appears better suited than $U(\bm{x})$. Indeed, the learning function $EFF(\bm{x})$ tends to explore more, whereas the function $U(\bm{x})$ focuses on a very accurate learning of the currently known limit state before exploring the rest of the domain. 
In the case of series system with disjoint failure domains, the initial GP can possibly not detect all the failure domains and the function $U(\bm{x})$ will learn very precisely the known branches before identifying the remaining ones, as presented in Sec. 4.1.2. of the article \cite{echard_ak-mcs:_2011}. It is thus possible that the probability of failure mean differs only very slightly for different GP trajectories when the GP prediction is very accurate on some limit state branches and has not detected that there are other failure regions. This is obviously counterproductive for the proposed algorithm. Indeed, we focus here on the global error on the probability of failure while the learning criterion based on $U$ is more concerned with the local error of the GP prediction.

The $EFF(\bm{x})$ learning function allows to avoid biased variance estimation as it is constructed to alternate more between exploration and exploitation during learning and thus allows to identify quicker the different branches of the limit state.

\subsection{Applications}
\label{sec:appli}
\subsubsection{Methodology settings and comparison measures}

It has been observed in many numerical applications that the squared-exponential kernel, also called Gaussian correlation model, is likely to undergo ill-conditioning  \cite{zimmermann_condition_2015}. In fact, the condition number depends on the number of sample points and the maximum distance between them. The covariance matrices built with this kernel are then particularly ill-conditioned when the training data are strongly correlated \cite{kostinski_condition_2000}, which is the case when an active learning algorithm is used for GP based reliability analysis.
 For this reason, in this paper the Mat\'{e}rn $\nicefrac{5}{2}$ kernel (see Eq.~\eqref{eq:kern_mat52}) will be used.
 
 Moreover, as for the AK methods, a maximum allowed COV has to be set as a stopping criterion of the algorithm. However, the COV computed in the proposed algorithm includes all uncertainties (not only the MC one) and the method is built to have balanced amount of variability due to both sources of uncertainty. This must be taken in consideration when choosing the value of $COV_{max}$.

The different active learning methods performances comparison can be based on different criterion or error measures such as:
\begin{itemize}
    \item $COV(\hat{P}_f)$: the COV of $\hat{P}_f$ estimations obtained on  $n_{run}$ runs of the estimation procedure of $\hat{P}_f$;
    \item{$e_r$: the mean over the the $n_{run}$ values of the absolute relative error between the estimations $\hat{P}_{f_i}, i=1, \ldots,n_{run}$ obtained and a reference value $P_{f_{ref}}$ (obtained e.g. by MCS with a very large number of samples)
    \begin{equation}
    e_r = \frac{1}{n_{run}} \sum_{i=1}^{n_{run}} \frac{|\hat{P}_{f_i}-P_{f_{ref}}|}{P_{f_{ref}}}
    \end{equation}
    }
    \item{ $\nu_{MC}$: a coefficient allowing to compare the numerical efficiency of the considered method to a classical MCS method, that is defined as follows: 
    \begin{equation}
        \nu_{MC} = \frac{N_{call}^{MC}}{N_{call}}
    \end{equation}
    where $N_{call}$ corresponds to the number of calls of the active learning method to the performance function to reach a COV equal to $COV(\hat{P}_f)$ and $N_{call}^{MC}$ is the number of samples needed by a MCS method (estimated by Eq.~\eqref{eq:COV_mc}) to obtain the same COV of $COV(\hat{P}_f)$ on the probability of failure.
    
    The efficiency $\nu_{MC}$ actually corresponds to the factor dividing the MCS budget to reach the same level of accuracy on ${P}_f$ with the active learning based method considered.
  }
\end{itemize}

 \subsubsection{Series system with four branches}\label{sec:appli_4branches_2}
We have applied the classical AK-MCS and the proposed methods on the example of a series system with four branches already defined in Sec.~\ref{sec:appli_4branches_1}.

\begin{figure}
\centering
\begin{subfigure}{\linewidth}
  \centering
  \includegraphics[width=1.\linewidth]{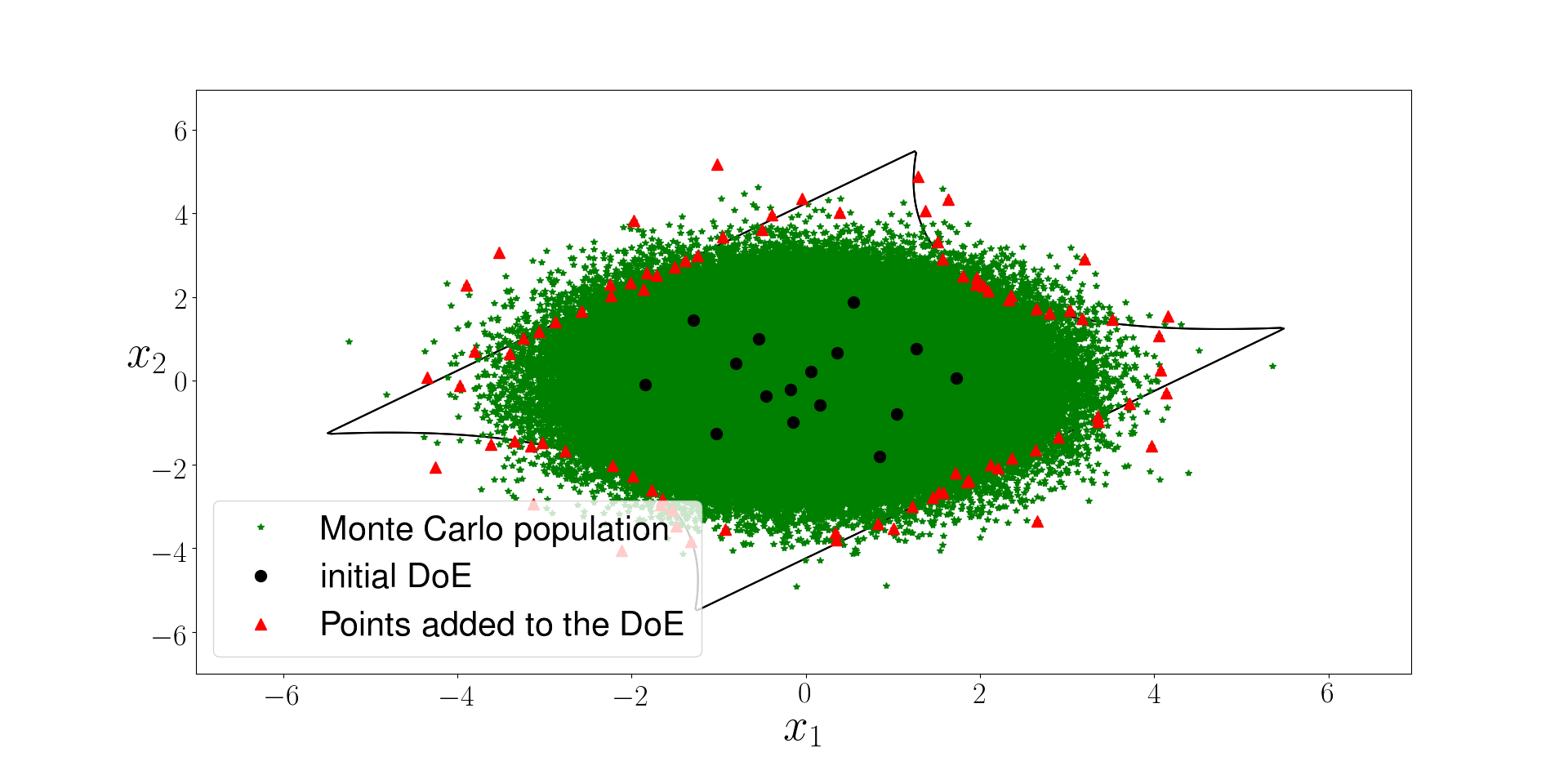}
  \caption{AK-MCS+$EFF$ result}
  \label{fig:4b_doe_akmcs}
\end{subfigure}\\
\begin{subfigure}{\linewidth}
  \centering
  \includegraphics[width=1.\linewidth]{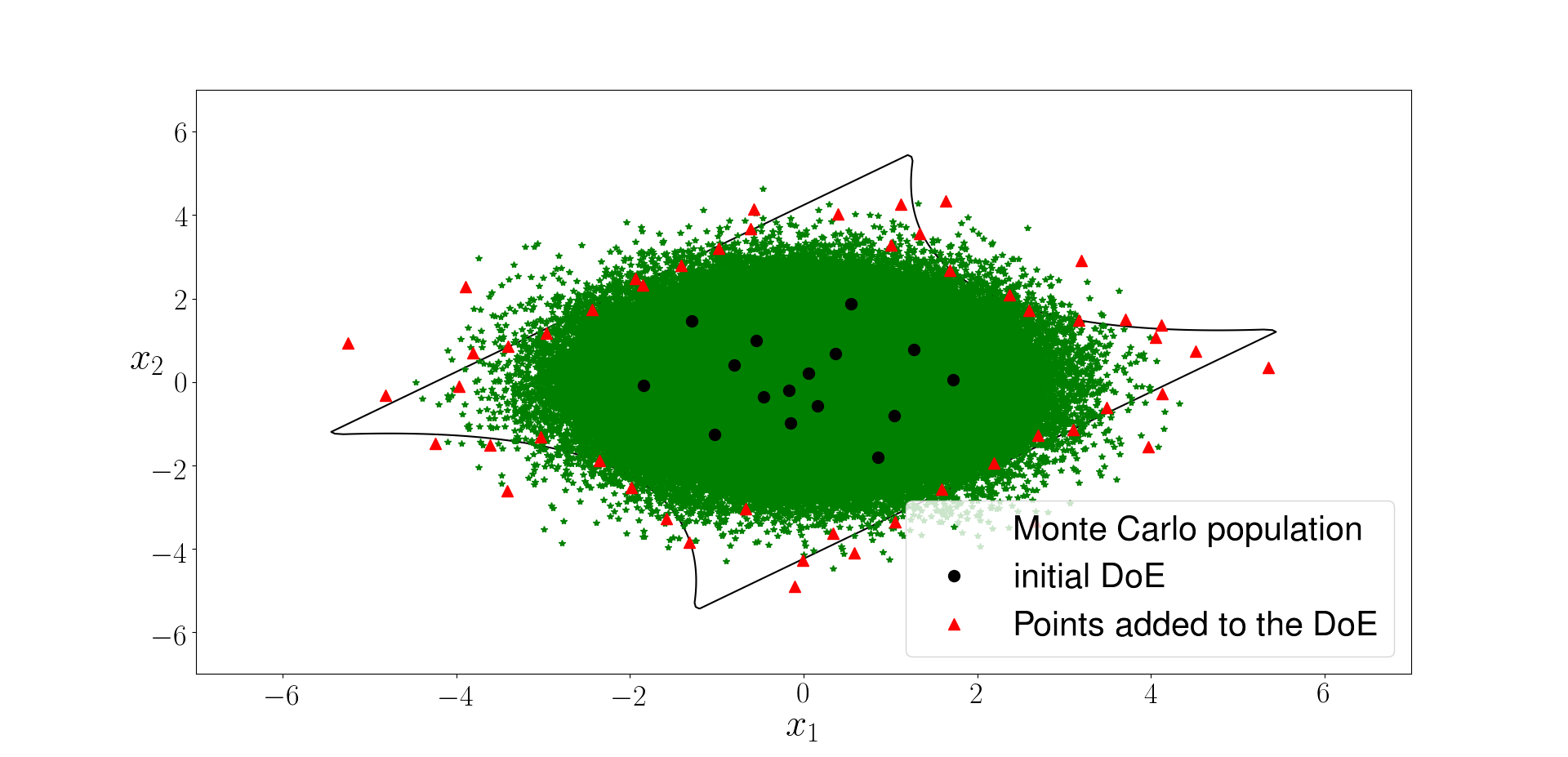}
  \caption{Vb-AGP + MCS approach result}
  \label{fig:4b_doe_gpmc}
\end{subfigure}
\caption{Comparisons of two DoEs: a) resulting from a run of AK-MCS+$EFF$ and b) one obtained with a run of the proposed method for the same inital DoE and MC population with $COV_{max}$ set to $3\%$ on the series system with four branches example}
\label{fig:4b_does_2methods}
\end{figure}

The proposed method was run for an initial DoE of 16 samples, an initial MC population of $5\times 10^4$ and a maximum coefficient of variation of $3\%$.
The final DoE resulting from a run of AK-MCS+$EFF$ and the final DoE obtained with a run of the proposed variance based algorithm on the same initial DoE and MC population are illustrated respectively on Fig.~\ref{fig:4b_doe_akmcs} and Fig.~\ref{fig:4b_doe_gpmc}. It can already be observed on these figures that the proposed variance based algorithm adds less points to the DoE to fulfill the learning stopping criterion.

\begin{figure}[h!]
    \centering
    \includegraphics[width=\textwidth]{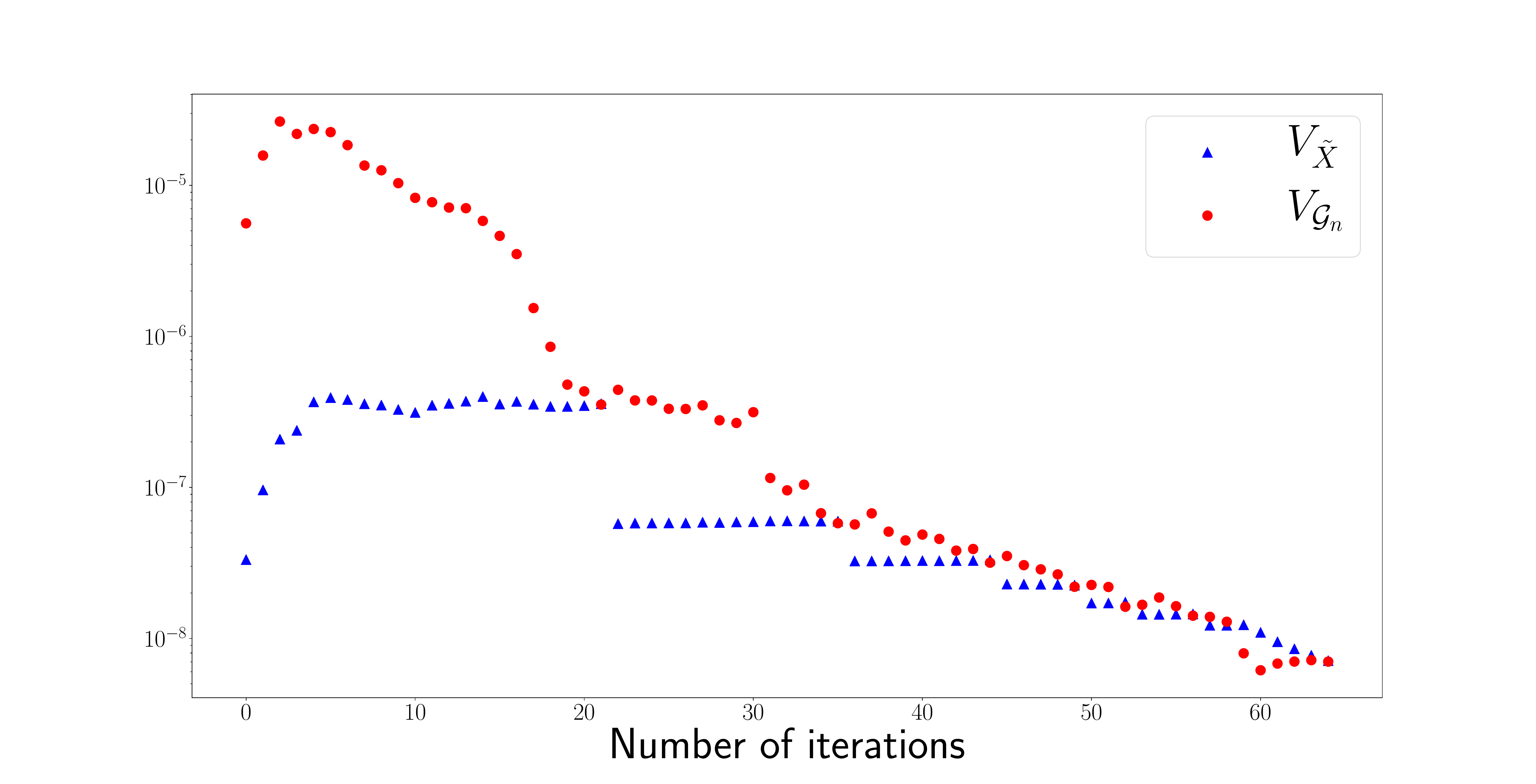}
    \caption{Evolution of the variance $V_{\tilde{\bm{X}}}$ and $V_{\mathcal{G}_n}$ estimations as a function of the number of iterations throughout a run of the proposed algorithm on the four branches test case}
    \label{fig:algomc_4b_var}
\end{figure}

The evolutions of both variance $V_{\tilde{\bm{X}}}$ and $V_{\mathcal{G}_n}$ estimations during a run are given in Fig.~\ref{fig:algomc_4b_var}.

 As for the algorithm AK-MCS, the influence of the GP is predominant at the beginning of the run and the surrogate model has to be enriched. However, we can see that at the end of the run the values of $V_{\tilde{\bm{X}}}$ and $V_{\mathcal{G}_n}$ are more balanced and $V_{\mathcal{G}_n}$ is not much lower than $V_{\tilde{\bm{X}}}$. Nonetheless, the estimated total COV at the end of the run is $2.9\%$ and respects thus the allowed COV of $3\%$. Here, the quantity $\frac{\sqrt{\widehat{V_{\tilde{\bm{X}}}}}}{\hat{P}_f}$ corresponding to the MC COV, that is usually used as a variability measure, is equal to $2.1\%$.

Then the algorithm was run 100 times with different initial DoEs of 16 points and initial Monte Carlo populations of $5\times 10^4$ and a maximum coefficient of variation of $3\%$ was set.
The reference result obtained, on average, by 100 runs of MCS ($n_{MC} =  10^{6}$) and the mean results of all methods are presented in Tab.~\ref{tab:results_4branches}.

\begin{table}[!h]
    \centering
    \begin{tabular}{*{7}{c}}
    \hline
            &&&&&&\\[-1.7ex] 
    Method & $N_{call}$&  $COV(N_{call})$ & $\hat{P}_f$ & $COV(\hat{P}_f)$ & $e_r$ & $\nu_{MC}$ \\[0.4ex] \hline 
        &&&&&&\\[-1.6ex] 
          MCS & $ 10^{6}$ &- & $4.46 \times 10 ^{-3}$ & $1.6\%$  & & \\[0.4ex]
          AK-MCS + $U$ & $128$ & $6.6\%$ &$4.48\times 10 ^{-3}$ & $3\%$ &  $2.4\%$ &  1976 \\[0.4ex]
            AK-MCS + $EFF$  &  $144$ & $6.6\%$ &$4.46\times 10 ^{-3}$ & $3\%$ &  $2.5\%$ & 1690   \\[0.4ex]
          Vb-AGP + MCS &  $68$ & $9.0\%$  & $4.46\times 10 ^{-3}$ &  $2.6\%$ &  $2.0 \%$ & 5146\\
    \end{tabular}
    \caption{Series system with four branches— 100 run mean results - Adaptive method parameters: $n_{MC}^{init}=5 \times 10^4$,  $n_{DoE}^{init}=16$, $COV_{max}=0.03$}
    \label{tab:results_4branches}
\end{table}

The results show that the method Vb-AGP allows to reduce the number of learning points needed to respect the same accuracy. Moreover, we can see that the variance (respectively COV) estimator proposed in this work is consistent with the empirical variance obtained on 100 runs of the algorithm. Indeed, the maximal imposed COV value is $3\%$ and an empirical COV of $2.6\%$ is obtained. 
 The mean ratios between the number of calls to the performance function by the Vb-AGP method and AK-MCS+$U$ and by the Vb-AGP method and AK-MCS+$EFF$ are respectively $1.9$ and $2.1$. Moreover, the value of the numerical efficiency indicator $\nu_{MC}$ of the method is 3 times higher than for AK-MCS+$EFF$ and 2.6 times higher than for AK-MCS+$U$.

\subsubsection{Dynamic response of a non--linear oscillator}  \label{sec:oscillo}

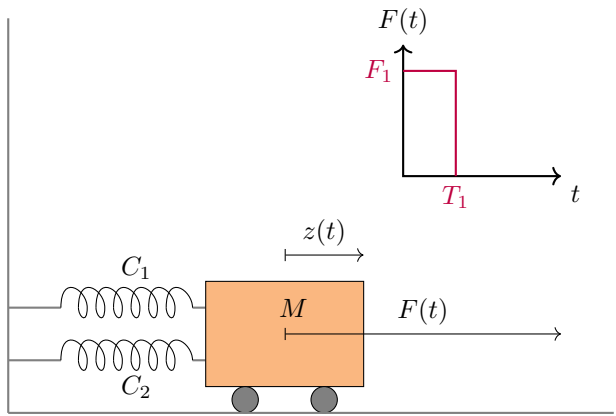
\begin{figure}[h!]
\centering
    \begin{tikzpicture}[scale=0.7, decoration={coil,aspect=0.4,segment length=3mm,amplitude=3mm}]
\tikzset{ressort/.style={thick,gray,smooth}}

 \draw [thick, gray] (0.5,2)--(1.5,2);
  \draw [thick, gray] (0.5,1)--(1.5,1);
   \draw [thick, gray] (4,2)--(4.25,2);
  \draw [thick, gray] (4,1)--(4.25,1);
\begin{scope}[shift={(1.5,2)},rotate=90]
 \foreach \r in {0,...,5}
 {
  \draw[scale=1/3,shift={(0,-\r)}]
	(0,0) .. controls ++(2,0) and ++(1,0) ..
	++(0,-1.5) .. controls ++(-1,0) and ++(-0.5,0) ..
	++(0,0.5);
  \draw[scale=1/3,shift={(-3,-\r)}]
	(0,0) .. controls ++(2,0) and ++(1,0) ..
	++(0,-1.5) .. controls ++(-1,0) and ++(-0.5,0) ..
	++(0,0.5);
 }
  \draw[scale=1/3,shift={(0,-6)}] (0,0) .. controls ++(2,0) and ++(1,0) .. ++(0,-1.5);
  \draw[scale=1/3,shift={(-3,-6)}] (0,0) .. controls ++(2,0) and ++(1,0) .. ++(0,-1.5);

\end{scope}
 \draw [fill= Apricot]  (4.25,0.5) rectangle (7.25,2.5)  ;
 \draw [fill=gray] (5,0.25) circle(0.25cm);
 \draw [fill=gray] (6.5,0.25) circle(0.25cm);

 \draw [thick, gray] (0.5,0)--(12,0);
  \draw [thick, gray] (0.5,0)--(0.5,7.5);
  
  \draw[|->] (5.75,1.5) -- (11,1.5) node [midway,above] {$F(t)$};
  \draw[|->] (5.75,3.) -- (7.25,3.) node [midway,above] {$z(t)$};
\node[text width=0.5cm] at (3,2.75) {$C_1$};
\node[text width=0.5cm] at (3,0.5) {$C_2$};
\node[text width=0.5cm] at (6,2) {$M$};

\draw [thick, <->]  (8,7) node [above] {$F(t)$}
-- (8,4.5) --  (11,4.5) node [below right] {$t$};
\draw [thick,purple, -]  (8,6.5) node [left] {$F_1$}
-- (9,6.5) --  (9,4.5)   node [below] {$T_1$} ;

\end{tikzpicture} 
    \caption{Non-linear oscillator}
    \label{fig:non_lin_osci}
\end{figure}

The example of non-linear oscillator is also widely used in the litterature and concerns the dynamic response of the nonlinear undamped single degree-of-freedom system illustrated in Fig.~\ref{fig:non_lin_osci}. This example is also studied in \cite{echard_kriging-based_2012,echard_combined_2013,lelievre_ak-mcsi:_2018}.

The corresponding performance function is expressed as: 
\begin{equation}
    G(C_1,C_2,M,R,T_1,F_1)=3R-\left|\frac{2F_1}{M\omega^2_0}\sin\left(\frac{\omega_0 T_1}{2}\right)\right|
\end{equation}
with $\omega_0=\sqrt{\nicefrac{(C_1+C_2)}{M}}$. 
Six random Variables listed in Tab.~\ref{tab:rand_oscillo} are considered for this problem. Actually, two cases are proposed here with a change of the applied force $F_1$ probability distribution parameters which lead to different probability of failure orders of magnitude.

\begin{table}[!h]
    \centering
    \begin{tabular}{*{4}{c}}
    \hline
         Variable & Distribution & Mean & standard deviation \\ \hline 
         $C_1$   &  Gaussian& 1 & 0.1 \\
        $C_2$   &  Gaussian& 0.1 & 0.01 \\
         $M$   &  Gaussian& 1 & 0.05 \\
          $R $  &  Gaussian& 0.5 & 0.05 \\
          $T_1 $  &  Gaussian& 1 & 0.2 \\
          $F_1$ -- Case 1 &  Gaussian& 1 & 0.2\\
            $F_1$ -- Case 2  &  Gaussian& 0.6 & 0.1 \\
    \end{tabular}
    \caption{Non-linear oscillator—random Variables}
    \label{tab:rand_oscillo}
\end{table}

The proposed method and the AK-MCS were applied on the first case with $F_1\sim \mathcal{N}(1,0.04)$ corresponding to a reference probability of failure of $2.86\times 10 ^{-2}$ (obtained with 100 runs of MCS with $n_{MC}=1 \times 10^{5}$). 
The methods were run 100 times for initial DoEs of 12 samples, initial MC populations of $1\times 10^4$ and a maximum coefficient of variation of $3\%$. The mean results are given in Tab.~\ref{tab:results_oscillo}.

\begin{table}[!h]
    \centering
     \begin{tabular}{*{7}{c}}
    \hline
                    &&&&&&\\[-1.7ex] 
    Method & $N_{call}$&  $COV(N_{call})$ & $\hat{P}_f$ & $COV(\hat{P}_f)$ & $e_r$ & $\nu_{MC}$ \\[0.4ex] \hline 
        &&&&&&\\[-1.6ex] 
         MCS & $1 \times 10^{5}$ &- & $2.86\times 10 ^{-2}$ & $2\%$ & - & -  \\[0.4ex]
          AK-MCS +$U$ & $59.8$ & $6.4\%$ &$2.85\times 10 ^{-2}$ & $2.7\%$ & $2.2\%$& 792 \\[0.4ex]
           AK-MCS +$EFF$  &  $52.5$ & $7.1\%$ &$2.87\times 10 ^{-2}$ & $2.8\%$ & $2.4\%$& 890 \\[0.4ex]
           Vb-AGP + MCS &  $22.5$ & $14.4\%$  & $2.84\times 10 ^{-2}$ &  $3.2\%$ &  $2.6\%$ & 1436 \\
    \end{tabular}
        \caption{Non-linear oscillator ($\mu_{F_1}=1$, $\sigma_{F_1}=0.2$) — 100 run mean results \newline Adaptive method parameters:  $n_{MC}^{init}=10^4$,  $n_{DoE}^{init}=12$, $COV_{max}=0.03$}
    \label{tab:results_oscillo} 
\end{table}


In terms of the numerical efficiency, the Vb-AGP method is also more effective than other techniques on this example as it provides the higher coefficient $\nu_{MC}$ value. Indeed, the coefficient $\nu_{MC}$ is 1.6 times higher than for AK-MCS + $U$ and 1.8 times higher than for AK-MCS + $EFF$. 

In Tab.~\ref{tab:results_oscillo}, the COV of $\hat{P}_f$ estimation is $3.2\%$ on the 100 runs of Vb-AGP and its $99\%$ confidence interval is given by $\left[ 0.022, 0.041 \right]$.

We can see that the COV of the number of calls $N_{call}$ is higher for the method Vb-AGP than for the other enrichment strategies. In this example the number of points added during the learning phase with the proposed method is of the same order of magnitude as the initial DoE. The results obtained are thus dependent on the initial DoE. We can suppose that the number of necessary enrichment points depends on the quality of the initial DoE, in terms of accuracy of the classification, and that is an underlying cause of the higher variance of $N_{call}$.

As presented previously, another case derived from the non linear oscillator example with another distribution of the variable $F_1$ can be achieved (Case 2 of Tab.~\ref{tab:rand_oscillo}). However, this test case corresponds to a very low probability of failure and can thus not be treated with the Vb-AGP + MCS method due to the limits of MCS use for rare events \cite{morio2014survey}.

Therefore, we propose another version of the method that involves Importance Sampling (IS) in order to address low probability of failure problems.

\section{Improvement with IS}
A way to highly decrease a Monte Carlo based estimator's variance is to use importance sampling instead of a classical Monte Carlo sampling. Moreover, the use of importance sampling allows to address rare event probabilities as the number of samples for the integration can be considerably reduced.

\subsection{Importance Sampling}\label{sec_IS}

The main idea of Importance Sampling (IS) is to find an auxiliary density $f_{aux}$, well-suited for the estimation of $P_f=\mathbbm{P}\left[G(\bm{x})\leq0\right]$, to generate \hbox{$n_{IS}\ll n_{MC}$} samples $\bm{X}_1,...,\bm{X}_{n_{IS}}\sim f_{aux}$ weighted for the estimation of the sought probability: 
\begin{equation}
\hat{P}_f^{IS}= \frac{1}{n_{IS}}\sum_{i=1}^{n_{IS}} w(\bm{X}_i) \mathbbm{1}_{G(\bm{X}_i)\leq 0}(\bm{X}_i)
\end{equation}
with $w(\bm{X}_i)=\frac{f_{\bm{X}}(\bm{X}_i)}{f_{aux}(\bm{X}_i)}$ the weight of the sample $\bm{X}_i$.
The auxiliary density $f_{aux}$ appears in the computation of the variance of the estimator in the following way:
\begin{equation}
   Var \left(\hat{P}_f^{IS}\right)=\frac{Var\left(w(\bm{X}_1) \mathbbm{1}_{G(\bm{X}_1)\leq 0}(\bm{X}_1)\right)}{n_{IS}}
  \label{eq:var_IS}
\end{equation}
while the variance of the MC estimator is $\frac{Var\left(\mathbbm{1}_{G(\bm{X}_1)\leq 0}(\bm{X}_1)\right)}{n_{MC}}$. Hence, if well chosen, the auxiliary density $f_{aux}$ can considerably reduce the variance of the estimator and improve its convergence. The best possible auxiliary density $f_{aux}$, denoted $f^{opt}_{aux}$, is the one that verifies $Var\left(\hat{P}_f^{IS}\right)=0$. Using Eq.~\eqref{eq:var_IS}, its expression is then given by \cite{bucklew2013introduction}:

\begin{equation}
f^{opt}_{aux}(\bm{x})=\frac{\mathbbm{1}_{G(x)\leq 0} f_{\bm{X}}(\bm{x})}{P_f}
\label{eq:faux_opt}
\end{equation}
Since it depends on the probability sought $P_f$ itself, it cannot be used directly. One of the difficulty of IS is then to compute an auxiliary density $f_{aux}$ as close as possible to $f^{opt}_{aux}$. Several methods have been developed such as \cite{de2005tutorial, zhang1996nonparametric,papaioannou2016sequential}. In this paper, we consider the non parametric adaptive IS (NAIS) proposed in \cite{morio_extremequantile} and detailed in \ref{sec:appendix_NAIS}. The principle of NAIS is to estimate iteratively the density $f^{opt}_{aux}$ with a weighted Gaussian kernel density. The advantage of this approach is its applicability on relatively complex failure domain as long as the dimensionality of the input is reasonable ($m<10$).

The probability of failure estimator obtained when running a GP active learning method combined with IS is given by:

\begin{equation}\label{eq:pf_gp_is}
\hat{P}_f(\bm{X}, \mathcal{G}_n)= \frac{1}{n_{IS}}\sum_{i=1}^{n_{IS}} w_i\mathbbm{1}_{\mathcal{G}_n(\bm{X}_i)\leq 0}(\bm{X}_i)
\end{equation}
with $\mathcal{G}_n$ the GP of $G$, $w_i=\frac{f_{\bm{X}}(\bm{X}_i)}{f_{aux}(\bm{X}_i)}$ the weights of the samples generated by IS and $f_{aux}$ the auxiliary IS density.

The effect of the GP and the IS accuracy on the probability of failure estimate can be obtained by rewriting the indices of variances proposed in Sec.~\ref{sec:var_estimators} adapted for IS.

\subsection{Variance based sensitivity index estimations}

The expected value of $\hat{P}_f$ knowing a population of $n_{IS}$ samples $\tilde{\bm{X}}=(\bm{X}_i)_{i=1,..,n_{IS}}$ generated by IS auxiliary density function $f_{aux}$ can be expressed  by rewriting Eq.~\eqref{eq:E_G_X} as follows:

\begin{align}
\begin{split}
  \mathbb{E}_{\mathcal{G}_n}\left[\hat{P}_f|\tilde{\bm{X}}\right] & =  \mathbbm{E}_{\mathcal{G}_n}\left[\frac{1}{n_{IS}}\sum_{i=1}^{n_{IS}} w_i\mathbbm{1}_{\mathcal{G}_n(\bm{X}_i)\leq 0}(\bm{X}_i)|(\bm{X}_i)_{i=1,..,n_{IS}}\right] \\
      & = \frac{1}{n_{IS}}\sum_{i=1}^{n_{IS}}w_i \mathbbm{E}_{\mathcal{G}_n}\left[\mathcal{B}(p(\bm{X}_i))\right] \\
   & = \frac{1}{n_{IS}}\sum_{i=1}^{n_{IS}}w_i p(\bm{X}_i)
   \end{split}
\end{align}
   
Hence, the variance $V_{\tilde{\bm{X}}}$ estimator is then obtained by adapting Eq.~\eqref{eq:V_X_est} for IS: 

\begin{align}
\begin{split}
        \widehat{V_{\tilde{\bm{X}}}}&= \frac{\widehat{Var_{\tilde{\bm{X}}}}((\bm{w}(\bm{\tilde{\bm{x}}}) \bm{p}(\tilde{\bm{x}}))}{n_{IS}}\\
        &=\frac{1}{n_{IS}(n_{IS}-1)}\sum_{i=1}^{n_{IS}}\left(w(\bm{X}_i)p(\bm{X}_i)-\frac{1}{n_{IS}} \sum_{j=1}^{n_{IS}} w(\bm{X}_j)p(\bm{X}_j)\right)^2 \label{eq:V_X_IS_est}
   \end{split}
\end{align}
and its $1-\alpha$ confidence interval estimated bounds can be expressed using the operators defined in Sec.~\ref{sec:theorie_var_int} and are given by:
\begin{equation}
\begin{array}{rcr}
      \widehat{V_{\tilde{\bm{X}}}^{inf}} &= & \frac{\widehat{Var_{\tilde{\bm{X}}}^{inf}}(\bm{w}(\bm{\tilde{\bm{x}}}) \bm{p}(\tilde{\bm{x}}))}{n_{IS}}\\
      \widehat{V_{\tilde{\bm{X}}}^{sup}}&  =& \frac{\widehat{Var_{\tilde{\bm{X}}}^{sup}}(\bm{w}(\bm{\tilde{\bm{x}}}) \bm{p}(\tilde{\bm{x}}))}{n_{IS}}
\end{array}    
\end{equation}

 As mentioned in Sec.~\ref{sec:var_estimators}, once the first term $V_{\tilde{\bm{X}}}$ is computed, the idea now is to express the $V_{\mathcal{G}_n}$ estimator.

The computation of the expected value of $\hat{P}_f$ knowing a realization of $\mathcal{G}_n$ can here be interpreted as a classical IS estimation for a deterministic model. Hence it follows this equality: 

\begin{align}
  \mathbb{E}_{\tilde{\bm{X}}}\left[\hat{P}_f|\mathcal{G}_n \right] & = 
   \mathbb{E}_{\tilde{\bm{X}}}\left[\frac{1}{n_{IS}}\sum_{i=1}^{n_{IS}} w_i \mathbbm{1}_{\mathcal{G}_n(\bm{X}_i) \leq 0}(\bm{X}_i)|\mathcal{G}_n\right]\\
   & = P_f(\mathcal{G}_n)
\end{align}

The probability of failure $P_f(\mathcal{G}_n)$ is here approached by an estimation by IS $$\hat{P}_f^{IS}(\mathcal{G}_n) = \frac{1}{n_{IS}}\sum_{i=1}^{n_{IS}} w_i \mathbbm{1}_{\mathcal{G}_n(\bm{x}_i) \leq 0}(\bm{x}_i),$$ with $(x_i)_{i=1,\ldots,n_{IS}}$ the samples of the IS population realization. Hence, as for MCS $V_{\mathcal{G}_n}$ can be numerically obtained by computing the IS estimator of $P_f$ for different trajectories of $\mathcal{G}_n$. The expression of $V_{\mathcal{G}_n}$ estimator given by Eq.~\eqref{eq:V_Gn_est} for MCS becomes: 
\begin{equation}\label{eq:V_Gn_IS}
     \widehat{V_{\mathcal{G}_n}}=\frac{1}{n_t-1} \sum_{i=1}^{n_t} \left(  \hat{P}_f^{IS}(G_i)  -\frac{1}{n_t} \sum_{j=1}^{n_t}  \hat{P}_f^{IS}(G_j) \right)^2 
\end{equation}
where $n_t$ is the number of $\mathcal{G}_n$ realizations.

Moreover its $1-\alpha$ confidence interval estimated bounds are given by: 
\begin{equation}
\begin{array}{rcr}
        \widehat{V_{\mathcal{G}_n}^{inf}}&  =& \widehat{Var_{\mathcal{G}_n}^{inf}}(\hat{P}_f^{IS}(\mathcal{G}_n))\\
       \widehat{V_{\mathcal{G}_n}^{sup}}&  =& \widehat{Var_{\mathcal{G}_n}^{sup}}(\hat{P}_f^{IS}(\mathcal{G}_n))
\end{array}    
\end{equation}

Finally, the probability of failure can be estimated by $\hat{P}_f^t$ given by Eq.~\eqref{eq:pf_final0} and the total variance can be estimated using the estimator given by Eq.~\eqref{eq:v_tot_est}, that are both computed using $\hat{P}_f$ values for $n_t$ realizations $(G_i, \tilde{\bm{X}}_i)$,\hbox{$\; i=1,\ldots,n_{t}$} are $n_{t}$  of $\mathcal{G}_n$ and IS population $\tilde{\bm{X}}$. 

Then the estimated total COV of an estimation $\hat{P}_f^{IS}$ of the probability of failure $\hat{P}_f$ obtained on an IS realization is given by: 
\begin{equation}
     \widehat{COV_{tot}}=\frac{\sqrt{\widehat{V_{tot}}}}{ \hat{P}_f^t}
     \label{eq:cov_tot_IS_est}
\end{equation}


\subsection{Extended method to Importance Sampling}

In order to address low probability of failure estimation problems, we propose to integrate IS to Vb-AGP.
The main idea is to replace the Monte Carlo population by an IS population for the probability of failure estimation. Thus, the variances $V_{\tilde{\bm{X}}}$ and $V_{\mathcal{G}_n}$ are obviously computed on the samples generated by IS $\tilde{\bm{X}}$ only and their estimations are obtained by applying Eq.~\eqref{eq:V_X_IS_est} and Eq.~\eqref{eq:V_Gn_IS}. However, the learning point candidates for GP improvement correspond to all samples generated throughout the run of the NAIS algorithm $\bm{X^{aux}}$ for the current auxiliary density function construction.

At the beginning of the algorithm, the initial learning points candidates are simply the samples of a classic Monte Carlo population generated with the distribution $f_{\bm{X}}$. Hence, the first probability of failure estimation is obtained using the MC estimator. Naturally, as soon as an IS population is used instead of the MC, the probability of failure estimator $\hat{P}_f$ is replaced by the one corresponding to IS given by Eq.~\eqref{eq:pf_gp_is}.

Then steps 8 and 9 described in Sec.~\ref{sec:new_method} for MCS are thus modified accordingly: 

\begin{description}
\item[8.]  If $\widehat{V_{\mathcal{G}_n}}   < \widehat{V_{\tilde{\bm{X}}}}$ then: 
\begin{description}
    \item[8.1.] If it is the first time the algorithm passes through this loop or when $\mathcal{G}_n$ has been updated, then a new auxiliary density function $f_{aux}$ is built with the GP $\mathcal{G}_n$. IS population and candidate samples for the learning $\bm{X^{aux}}$ are also replaced by the most recent ones generated. Then the algorithm goes back to step 4. 
    
     Otherwise the algorithm goes to step 8.2.
    \item[8.2.] New samples are added to the IS population and the method goes back to step 4. 
\end{description}
Else if  $\widehat{V_{\mathcal{G}_n}}  > \widehat{V_{\tilde{\bm{X}}}}$, the algorithm goes to step 9.
    \item[9.] The learning function $EFF(\bm{x})$ is evaluated on the whole samples candidate population $\bm{X^{aux}}$ to find the best candidate $x^*$ to evaluate for enriching the GP metamodel. The performance function is evaluated on the sample $x^*$ and the DoE is enriched with this new observation. Then the method goes to step 3 to update the GP model. 
\end{description}

\begin{figure}[h!]
    \centering
    \includegraphics[width=\linewidth]{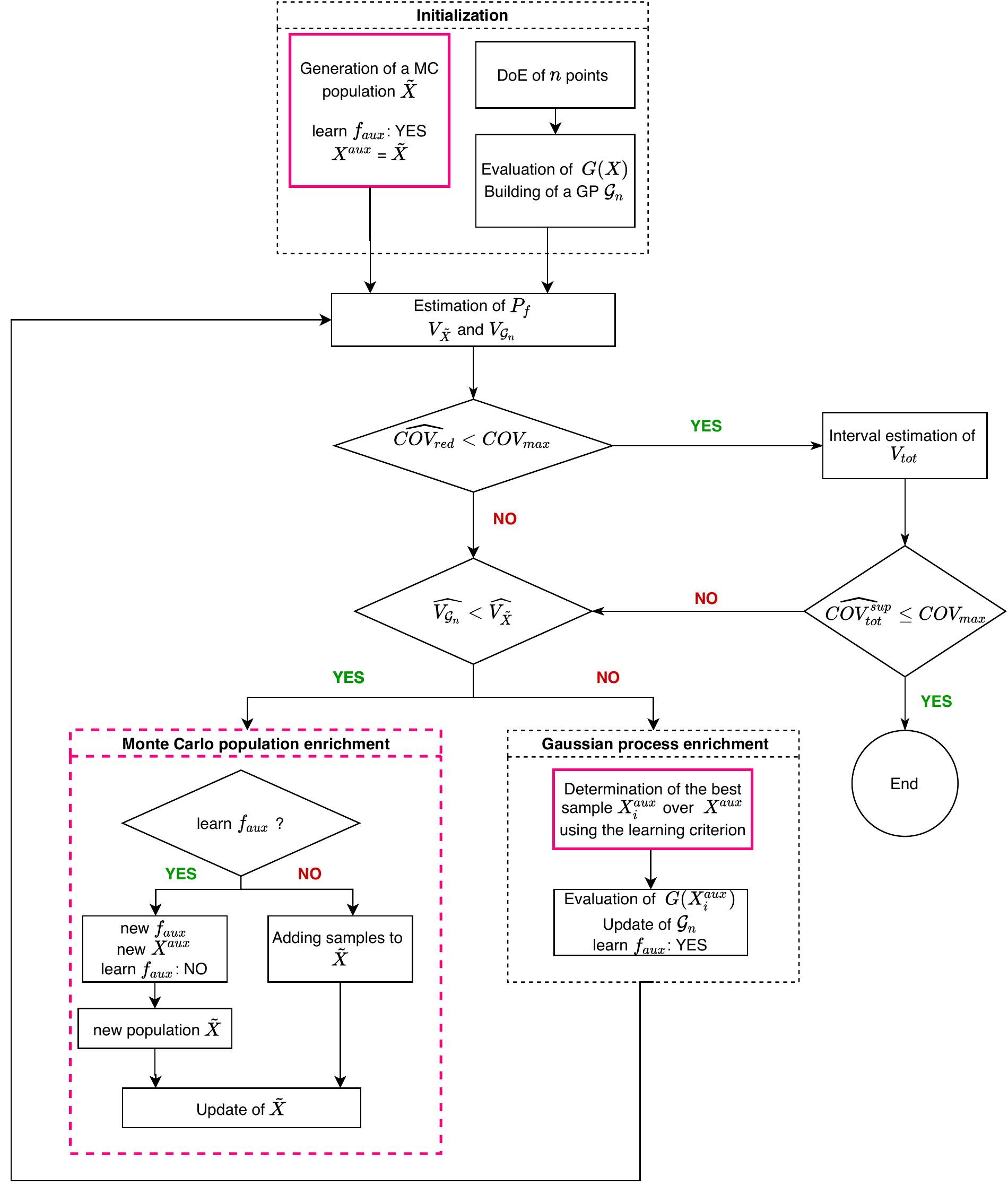}
    \caption{Flowchart of the learning algorithm improved with IS}
    \label{fig:sensitivity_is}
\end{figure}

The extended method procedure is summarized in Fig.~\ref{fig:sensitivity_is}. 

Moreover, as the extended method with IS may address low probability problems, there are great chances that the probability estimated with the initial DoE and MC population is equal to zero. In order to
 address this problem, NAIS is run one time on the GP built with the initial DoE. That allows to estimate a first auxiliary density function but also more appropriate candidate samples for the GP learning. 

Indeed, for very low probability of failure a suited initial DoE to have an appropriate initial GP approximation to continue the learning should be sampled in a certain vicinity of the limit state. However, due to the lack of information on the failure domain at the algorithm initialisation the first estimated auxiliary density function for IS may not actually corresponds to the optimal one. Therefore, a second initial DoE of size $2m$ more relevant for the learning is generated after the run of NAIS. 
The points of the DoE are chosen by an iterative selection by using the $EFF$ learning criterion on all intermediate samples generated throughout the NAIS run, with an update of the GP after each point added to the DoE.

\subsection{Applications}

\subsubsection{Low probability series system with four branches}

\begin{figure}
    \centering
    \includegraphics[width=1.2\linewidth]{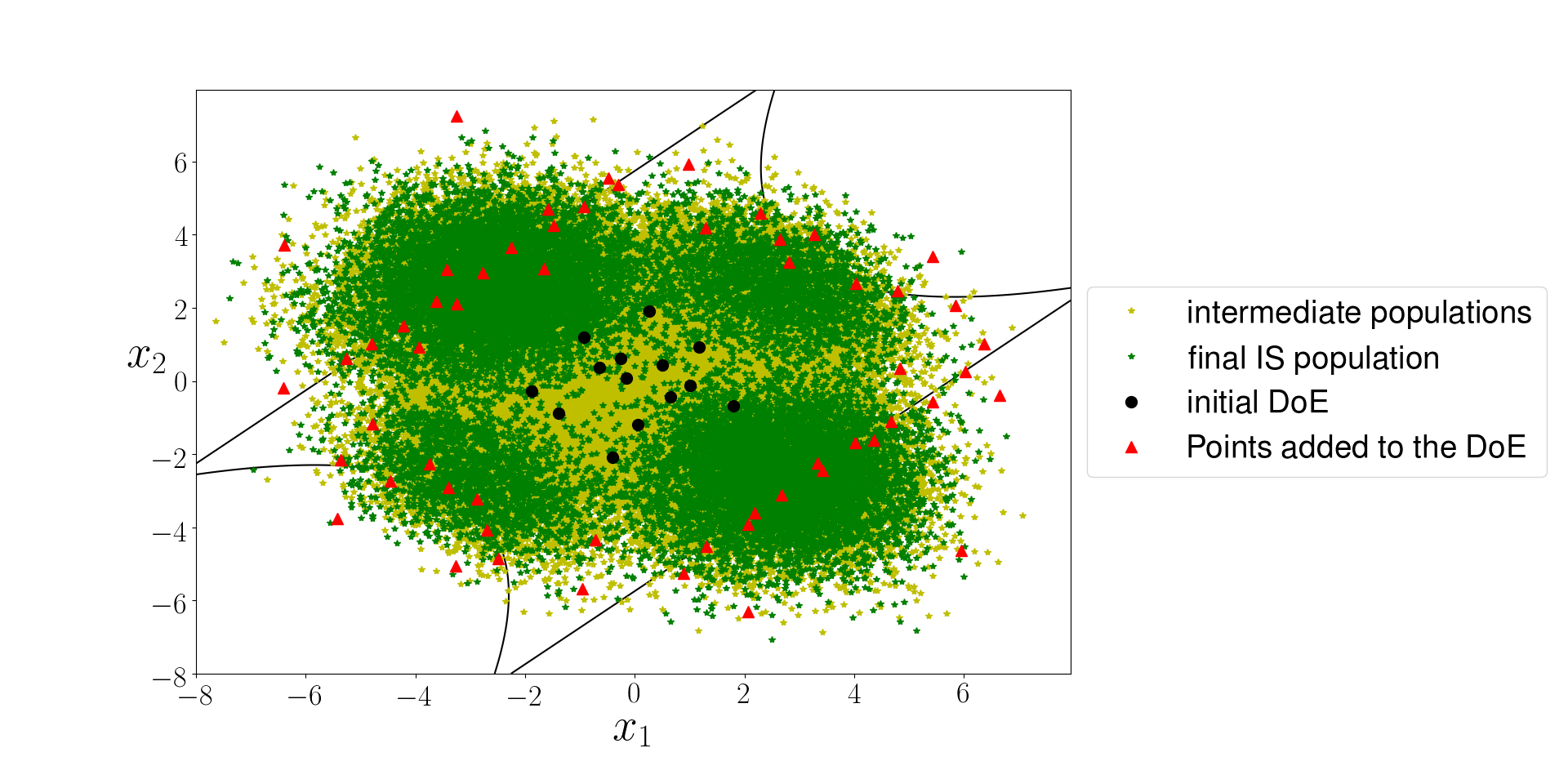}
    \caption{Vb-AGP + IS (NAIS) method}
    \label{fig:4b_low}
\end{figure}

The Vb-AGP + IS method was applied on a test case derived from the series system with four branches limit state function $G(\bm{x})$ defined by Eq.~\eqref{eq:4b_limitstate}. The failure is defined here by $G(\bm{x}) \leq 1.5$ and the related reliability problem corresponds to a probability of failure of $5.29 \times 10 ^{-5}$ with a COV $2.1\%$ estimated by MCS (100 runs for $n_{MC} = 5 \times 10^7$).
The proposed variance based method Vb-AGP with the adaptive IS method NAIS applied on this test case is illustrated on Fig.~\ref{fig:4b_low}. On this Figure, the intermediate population used as GP learning samples and the IS population used to compute the probability are represented.

The mean results over 100 runs of the algorithm on this test case are given in Tab.~\ref{tab:4b_low}. They show that the Vb-AGP + IS method allows to respect the maximal COV set to $3\%$ on the 100 runs of the algorithm.

\begin{table}[!h]
    \centering
    \begin{tabular}{*{7}{c}}
    \hline
                    &&&&&&\\[-1.7ex] 
    Method & $N_{call}$&  $COV(N_{call})$ & $\hat{P}_f$ & $COV(\hat{P}_f)$ & $e_r$ & $\nu_{MC}$ \\[0.4ex] \hline 
        &&&&&&\\[-1.6ex] 
          MCS &  $5 \times 10^{7}$ & - & $5.29 \times 10 ^{-5}$ & $2.1\%$ & -  & -  \\[0.4ex]
          Vb-AGP + IS &  $104$ & $16.2\%$  & $5.33 \times 10 ^{-5}$ &  $3.0\%$ & $2.4\%$ & 201437 \\
    \end{tabular}
        \caption{Series system with four branches low probability of failure problem - 100 run mean results - Adaptive method parameters: $n_{DoE}^{init}=12$, $COV_{max}=0.03$}
    \label{tab:4b_low}
\end{table}

Here, we do not compare the method to AK-MCS because this type of reliability problem with low probabilities of failure is difficult to address with this method and neither to AK-IS \cite{echard_combined_2013} as the method is based on FORM and thus not suited for multimodal failure domains.  Nonetheless, it can be noted that the efficiency indicator $\nu_{MC}$ of the method in comparison to MCS is very high.

\subsubsection{Dynamic response of a non-linear oscillator}

The variance based GP+IS method was then applied on the second case derived from the example dealing with the dynamic response of a non-linear oscillator, described in Sec.~\ref{sec:oscillo}, with $F_1\sim \mathcal{N}(0.6,0.01)$ corresponding to an estimated probability of failure of $9.08\times 10 ^{-6}$ ($2.47\%$) with MCS. 
The method was run 50 times for initial DoEs of 12 samples and a maximum coefficient of variation of $3\%$.

Obviously, this reliability problem is intractable with the AK-MCS method but has been handled by the method AK-IS \cite{echard_combined_2013}. Hence, we compare the mean results of the Vb-AGP + IS approach with the results of AK-IS presented in the paper \cite{echard_combined_2013}. These results are given in Tab.~\ref{tab:results_oscillo_low}.

    \begin{table}[!h]
    \centering
    \begin{tabular}{*{7}{c}}
    \hline
    &&&&&&\\[-1.7ex] 
    Method & $N_{call}$&  $COV(N_{call})$ & $\hat{P}_f$ & $COV(\hat{P}_f)$ & $e_r$ & $\nu_{MC}$ \\[0.4ex] \hline 
        &&&&&&\\[-1.6ex] 
     MCS & $1.8 \times 10^{9}$ &- & $9.08 \times 10 ^{-6}$ & $2.47\%$ & &  \\[0.4ex]
          FORM+ IS  & $29 + 10^{4}$ &- & $9.13 \times 10 ^{-6}$ & $2.29 \%$ & &  \\[0.4ex]
           AK-IS \citep{echard_combined_2013} & $29+ 38$ & - & $9.13 \times 10 ^{-6}$ & $2.29 \%$ & &  1943812 \\[0.4ex]
            Vb-AGP  + IS  &  $58$ & $20\%$  & $9.09\times 10 ^{-6}$ &  $2.9\%$ & $2.3\%$ & 2240946  \\
    \end{tabular}
        \caption{Non--linear oscillator ($\mu_{F_1}=0.6, \sigma_{F_1}=0.1$) — 100 run mean results \newline Adaptive method parameters: $n_{DoE}^{init}=12$, $COV_{max}=0.03$}
    \label{tab:results_oscillo_low}
\end{table}
On this example, the method Vb-AGP divides on average the number of calls to the performance function by AK-IS by a factor 1.16. This reduction is less important than the ones obtained on the previous examples with AK-MCS. Note that the COV of $N_{call}$ is relatively high for the method Vb-AGP+IS. That can be explained by the influence of the initial DoE, that has an important impact on the performances of NAIS. Moreover, six random variables are considered on this example and thus the efficiency of NAIS is considerably reduced as the stochastic dimension is quite high for the applicability of this method.

\section{Conclusions}
In this paper, we showed that the effect of both the Monte Carlo sampling and the GP surrogate model on the probability of failure estimator can be analyzed by a sensitivity analysis based on variance decomposition. Then we have proposed estimators of the variances related to each of these two uncertainty sources and also an estimator of the total variance in order to compute them numerically. This analysis enables us to quantify the source of uncertainty that has the most impact on the variability of the probability of failure estimation. 

Then we proposed a Variance based Active GP (Vb-AGP) learning procedure that integrates this analysis to improve the major source of uncertainty during the learning phase and a stopping criterion based on the total variance of the probability of failure estimation. The method was applied on two examples and showed great potential to reduce the number of learning points while satisfying the maximum COV constraint imposed by the user. Moreover, the method gives an estimation of the total COV that has been validated on both examples.

An extension of this method to IS, in order to make it more suitable to probability of rare events estimation, was then presented. In this work, the adaptive IS algorithm NAIS was used in order to tackle multimodal failure domains without any a priori knowledge of the domain. The approach was applied on two examples and was shown effective in terms of the accuracy of the probability of failure total COV estimations and also in terms of potential number of simulations reduction. 
However, it should be noted that the application of Vb-AGP + MCS is limited to problems with a relatively low stochastic dimension because of the memory cost of accurate trajectory simulations with the Karhunen-Loeve expansion for large MC populations. Hence, a potential improvement of the method could be thought by looking to improve the efficiency of the trajectory simulations technique while taking care that the trajectories approximations are sufficiently accurate. 

Note also that the use of the Vb-AGP + IS method is a good way to reduce the sampling population size, but as the NAIS algorithm is only efficient for problems with relatively low input dimensionality ($m<10$) it is also not suited for problems with high stochastic dimensions. 
In this work, an extension to rare events problems with NAIS was chosen as we considered that there is no a priori knowledge of the limit state. However, in case some hypothesis about the limit state form are available by experience for example, other IS or Subset Sampling techniques might be considered to circumvent the curse of dimensionality.

\section*{Acknowledgment}

This work was supported by the French National Research Agency (ANR) through the ReBReD project under grant ANR-16-CE10-0002.


\appendix
\section{Gaussian process trajectories computations by simulating an unconditioned Gaussian process}\label{appendix:gp_traj}
The objective here is to explain how to avoid numerical issues experienced when simulating GP trajectories with large population.

We saw in Sec.~\ref{sec:var_estimators} that the estimation of $V_{\mathcal{G}_n}$ can be assessed from realizations of the Gaussian process ${\mathcal{G}_n}$ at each point of the sampled population. 

Let $\mathcal{G}_n$ be a conditioned GP with mean function $\bm{\mu}_n(\cdot)$ and covariance matrix $\bm{C}_n(\cdot)$. A trajectory (or realization) of the GP $\mathcal{G}_n$ at the Monte Carlo population $\tilde{\bm{X}}$ can actually be expressed as follows: 
\begin{equation}
  \mathcal{G}_n(\tilde{\bm{X}})=  \bm{\mu}_n(\tilde{\bm{X}}) + \bm{L}_n(\tilde{\bm{X}}) \bm{\xi}
\end{equation}
with $\bm{\xi} \sim \mathcal{N}(\bm{0}_n,\bm{I}_n)$ and $\bm{L}_n(\tilde{\bm{X}}) \in M^{n_{MC}}(\mathbbm{R})$ the Cholesky factorization matrix of $\bm{C}_n(\tilde{\bm{X}})$, i.e. $\bm{C}_n(\tilde{\bm{X}})= \bm{L}_n(\tilde{\bm{X}})\bm{L}_n(\tilde{\bm{X}})^T$.


However for large populations, the computation of $\mathcal{G}_n(\tilde{\bm{X}})$ realizations present huge numerical costs or encounter numerical issues like ill-conditioned covariance matrix, making it impossible to compute the Cholesky decomposition and thus the GP realizations by this method. 
These numerical problems can be avoided by simulating an unconditioned Gaussian process \cite{rasmussen_gaussian_2006,villemonteix_informational_2009,le_gratiet_bayesian_2014}. 

Let $\tilde{\mathcal{G}}(\bm{x})$ be a centered Gaussian process with the same covariance function as $\mathcal{G}(\bm{x})$: 
\begin{equation}
\tilde{\mathcal{G}}(\bm{x}) \sim GP(0,\sigma^2_Z r(\bm{x},\bm{x'}))
\end{equation}
 and let $\tilde{\mu}(\bm{x}) $ be its prediction mean based on the random variables $\tilde{\mathcal{G}}(\bm{x_{doe}})$.
 
Then, let us define the Gaussian process $\tilde{\mathcal{G}}_n(\bm{x})$: 
\begin{equation}
\tilde{\mathcal{G}}_n(\bm{x}) = \mu_n(\bm{x})- \tilde{\mu}(\bm{x})+\tilde{\mathcal{G}}(\bm{x}) \label{eq:def_gn_approx}
\end{equation}
with $\mu_n(\bm{x})$ the mean of the prediction by $\mathcal{G}_n(\bm{x})$ at point $\bm{x}$ and: 
\begin{equation}
\tilde{\mu}(\bm{x}) = \bm{f}(\bm{x})^T\bm{\tilde{\beta}}+\bm{k}(\bm{x})^T\bm{C}^{-1}(\bm{y}-\bm{F}\bm{\tilde{\beta}})
\end{equation}
with $\bm{\tilde{\beta}}=(\bm{F}^T\bm{C}^{-1}\bm{F})^{-1}\bm{F}^T\bm{C}^{-1}\tilde{\mathcal{G}}(\bm{x_{doe}})$. 

Then, $\tilde{\mathcal{G}}_n(\bm{x})$ has the same distribution as $\mathcal{G}_n(\bm{x})$ conditionally to past observations $(\bm{x_{doe}},\bm{y})$. In other words, we have:
\begin{equation}
    \tilde{\mathcal{G}}_n(\bm{x})\stackrel{\mathcal{L}}{=} \mathcal{G}_n(\bm{x})
\end{equation}

Hence, a simulation of $\mathcal{G}_n(x)$ can be obtained by adding to its mean $\mu_n(\bm{x})$ the prediction error $\tilde{\mathcal{G}}(\bm{x})-\tilde{\mu}(\bm{x})$ of $\tilde{\mathcal{G}}(\bm{x})$. It allows to simulate the centered Gaussian process $\tilde{\mathcal{G}}(\bm{x})$ instead of $\mathcal{G}_n(x)$. Contrary to the conditioned Gaussian process, the unconditioned Gaussian process variance values at points in the vicinity of the DOE are not close to zero. It allows a better conditioning of the covariance matrix and thus avoids the related numerical issues. 

Moreover, it allows to use computationaly efficient representations of random fields such as the Karhunen-Loève (KL) expansion. The numerical approximation of the KL expansion can be obtained by using the Nyström procedure or Galerkin methods as presented in \cite{betz_numerical_2014}. 
Hence, once the KL decomposition of the Gaussian process $\tilde{\mathcal{G}}(\bm{x})$ is estimated it can be used to easily obtain realizations at any point $\bm{x}$ using Eq.~\eqref{eq:def_gn_approx}.

\section{Non parametric adaptive importance sampling (NAIS)} \label{sec:appendix_NAIS}

The goal of the non parametric adaptive importance sampling (NAIS) algorithm is to compute an estimator of the optimal auxiliary density function $f^{opt}_{aux}$, given by Eq.~\eqref{eq:faux_opt}, with standard Gaussian kernel density functions $K_d(\cdot)$ weighted with weights \text{w}, starting from the input density function $f_{\bm{X}}$. This method is well adapted to complex failure region $D_f$.

The different steps of the algorithm are described in Algorithm~\ref{algo:Nais}.

The successive NAIS iterations allow to update the estimator of the optimal auxiliary density function $\hat{g}_{opt}^{k}$. In the end we obtain the estimator of the optimal auxiliary density function of the initial sought probability $P_f=\mathbbm{P}[\mathcal{G}_n(\bm{X})\leq 0]$. The probability is then estimated with the IS formula (line 19 Algorithm~\ref{algo:Nais}).

\begin{algorithm}
\caption{Non parametric Adaptive Importance Sampling (NAIS)}
\begin{algorithmic}[1]  \label{algo:Nais}
\REQUIRE $\rho, n_{IS}, f_{\bm{X}}, \mathcal{G}_n$
\STATE $k \leftarrow 0$
\STATE $\tilde{\bm{X}}^{0} \leftarrow \text{population generated from } f_{\bm{X}} \text{ of cardinal }  n_{IS} $
\STATE $\textbf{Y}^{0} \leftarrow \mathcal{G}_n(\tilde{\bm{X}}^{0})$
\STATE $S^*\leftarrow \rho \text{-quantile of }\textbf{Y}^{0} $
\STATE $\gamma_0 \leftarrow$ max($S^*,0)$
\STATE $\textbf{w}^0 \leftarrow \mathbb{I}_{\textbf{Y}^{0}\leq\gamma_0} \hspace{2cm} $
\STATE $I_0 \leftarrow \frac{1}{n_{IS}} \sum_{i=1}^{n_{IS}} w_i^0$
\STATE $\hat{g}_{opt}^{1} (\bm{x}) \leftarrow \frac{1}{n_{IS} \text{det}(B_N) I_0}\sum_{i=1}^{n_{IS}} w_i^0 K_d\left(B_N^{-1}\left(\bm{x}-\bm{X}_i^0\right)\right)$ 
\WHILE{$\gamma_k > 0$}
\STATE $k \leftarrow k+1$
\STATE $\tilde{\bm{X}}^{k} \leftarrow \text{population generated from } \hat{g}_{opt}^{k} \text{ of cardinal } n_{IS} $
\STATE $\textbf{Y}^{k} \leftarrow \mathcal{G}_n(\tilde{\bm{X}}^{k})$
\STATE $S^*\leftarrow \rho \text{-quantile of }\textbf{Y}^{k} $
\STATE $\gamma_k \leftarrow$ max($S^*,0)$
\STATE $\text{w}_i^j \leftarrow \frac{\mathbbm{1}_{\textbf{Y}_i^{j}\leq\gamma_k}f_{\bm{X}}(\bm{X}_i^j)}{\hat{g}_{opt}^{j}(\bm{X}_i^j)}, \qquad \text{for $i,j \in  [1,n_{IS}]\times[1,k]$}$
\STATE $I_k \leftarrow \frac{1}{k n_{IS}} \sum_{j=1}^k \sum_{i=1}^{ n_{IS}} \text{w}_i^j$
\STATE $\hat{g}_{opt}^{k+1} (\bm{x}) \leftarrow \frac{1}{k n_{IS} \text{det}(B_N) I_k}\sum_{j=1}^k \sum_{i=1}^{ n_{IS}} \text{w}_i^j K_d\left(B_N^{-1}\left(\bm{x}-\bm{X}_i^j\right)\right)$ 
\ENDWHILE

\STATE $\hat{P}_f^{NAIS} \leftarrow \frac{1}{n_{IS}} \sum_{i=1}^{n_{IS}} \frac{\mathbb{I}_{\textbf{Y}^k\leq 0}f_{\bm{X}}(\bm{X}^k_i)}{\hat{g}_{opt}^k(\bm{X}^k_i)}$
\end{algorithmic}
\end{algorithm} 
In Algorithm~\ref{algo:Nais} line 8, $B_N$ is a diagonal covariance matrix.

In practice the GP prediction mean is used to evaluate the samples. However in the proposed method, the NAIS algorithm is used in intermediate steps were the GP approximation of the limit state may not be accurate. Hence, it is more appropriate to use an auxiliary density suited to estimate the probability $\hat{P}_f^{NAIS}= \frac{1}{n_{IS}} \sum_{i=1}^{n_{IS}} p(\bm{X}_i)\frac{f_{\bm{X}}(\bm{X}_i)}{\hat{g}_{opt}^k(\bm{X}_i)}$. 
Therefore in our method, the NAIS algorithm is run with weights computed as follows: 
\begin{equation}
    \text{w}_i^j = \frac{\mathbbm{P}[\mathcal{G}_n(\bm{X}_i^j)\leq \gamma_k]f_{\bm{X}}(\bm{X}_i^j)}{\hat{g}_{opt}^{j}(\bm{X}_i^j)}, \qquad \text{for $i,j \in  [1,n_{IS}]\times[1,k]$}
\end{equation}
with $\mathbbm{P}[\mathcal{G}_n(\bm{X})\leq \gamma_k]=\Phi\left(\frac{\gamma_k-\mu_n(\bm{x})}{\sigma_n(\bm{x})}\right)$

Moreover, in case of very low probability of failure the run of NAIS with the first built GPs can fail as the mean predicted values become quasi constant as the NAIS intermediate populations expand outward the known points. This is due to the initial large uncertainty on the failure domain. Therefore, a maximum relative residual between two consecutive intermediate thresholds $\gamma_k$ must be set. On our test cases, a relative residual of $10^{-3}$ has been set.

\end{document}